\pdfoutput=1

\documentclass[11pt]{article}

\usepackage[]{EMNLP2023}

\usepackage{times}
\usepackage{latexsym}

\usepackage[T1]{fontenc}

\usepackage[utf8]{inputenc}

\usepackage{microtype}

\usepackage{inconsolata}

\usepackage{amsmath,amssymb,amsthm}
\usepackage{multirow}           
\usepackage{booktabs}           
\usepackage{arydshln}
\usepackage{graphicx}           
\usepackage{float}              
\usepackage{bm}                 
\usepackage[tikz]{bclogo}       
\usepackage{amsmath}            
\usepackage{amsfonts}           
\usepackage{pgfplots}           
\pgfplotsset{compat=1.18}
\usepackage{caption}

\usepackage[hang,flushmargin]{footmisc}

\definecolor{brickred}{HTML}{b92622}
\definecolor{midnightblue}{HTML}{005c7f}
\definecolor{salmon}{HTML}{f1958d}
\definecolor{burntorange}{HTML}{f19249}
\definecolor{junglegreen}{HTML}{4dae9d}
\definecolor{forestgreen}{HTML}{499c5e}
\definecolor{pinegreen}{HTML}{3d8a75}
\definecolor{seagreen}{HTML}{6bc1a2}
\definecolor{limegreen}{HTML}{97c65a}

\title{MixEdit: Revisiting Data Augmentation and Beyond for\\Grammatical Error Correction}

\author{
Jingheng Ye$^{1}$,~Yinghui Li$^{1*}$,~Yangning Li$^{1,2}$,~Hai-Tao Zheng$^{1,2}$\thanks{$^{\dagger}$Corresponding authors: Hai-Tao Zheng and Yinghui Li.}\\
        $^{1}$Tsinghua Shenzhen International Graduate School, Tsinghua University \\ 
        $^{2}$Peng Cheng Laboratory \\
        \texttt{\{yejh22,liyinghu20\}@mails.tsinghua.edu.cn}
}

\begin{document}
\maketitle

\begin{abstract}

Data Augmentation through generating pseudo data has been proven effective in mitigating the challenge of data scarcity in the field of Grammatical Error Correction (GEC).
Various augmentation strategies have been widely explored, most of which are motivated by two heuristics, i.e., increasing the distribution similarity and diversity of pseudo data.
However, the underlying mechanism responsible for the effectiveness of these strategies remains poorly understood.
In this paper, we aim to clarify how data augmentation improves GEC models.
To this end, we introduce two interpretable and computationally efficient measures: Affinity and Diversity.
Our findings indicate that an excellent GEC data augmentation strategy characterized by high Affinity and appropriate Diversity can better improve the performance of GEC models.
Based on this observation, we propose MixEdit, a data augmentation approach that \textit{strategically} and \textit{dynamically} augments realistic data, without requiring extra monolingual corpora.
To verify the correctness of our findings and the effectiveness of the proposed MixEdit, we conduct experiments on mainstream English and Chinese GEC datasets.
The results show that MixEdit substantially improves GEC models and is complementary to traditional data augmentation methods~\footnote{
All the source codes of MixEdit are released at~\url{https://github.com/THUKElab/MixEdit}.
}.

\end{abstract}
\begin{table*}[tbp!]
\centering
\scalebox{0.75}{
\begin{tabular}{ll}
\toprule

\textbf{Target}  &
This will , if not already , cause problems as there is very limited space for us .     \\

\hdashline

\textbf{Source}  &
This will , if not already , \textcolor{red}{caused} problems as there \textcolor{red}{are} very limited \textcolor{red}{spaces} for us .   \\

\textbf{Direct Noise (DN)}  &
$\langle$mask$\rangle$ will , \textcolor{red}{I} $\langle$mask$\rangle$ already $\langle$mask$\rangle$ will cause problems \textcolor{red}{the} $\langle$mask$\rangle$ is $\langle$mask$\rangle$ $\langle$mask$\rangle$ space for $\langle$mask$\rangle$ .       \\

\textbf{Pattern Noise (PN)}  &
This will , if not already , cause problem as there is very \textcolor{red}{limiting} space for us \textcolor{red}{?}     \\

\textbf{Backtranslation (BT)}  &
This \textcolor{red}{if} not already \textcolor{red}{cause} the problems as there \textcolor{red}{are} very \textcolor{red}{few} space for us .             \\

\textbf{Round-translation (RT)}  &
 \textcolor{red}{If that had not been done} , it would have caused problems , because our space is very limited .     \\

\midrule

\textbf{Target}  &
We realize that burning of fuels produces a large amount of greenhouse gases .      \\

\hdashline

\textbf{Source}  &
We \textcolor{red}{relize} that burning of fuels \textcolor{red}{produce the} large amount of greenhouse gases .      \\

\textbf{Direct Noise (DN)}  &
We $\langle$mask$\rangle$ $\langle$mask$\rangle$ burning of \textcolor{red}{looking} $\langle$mask$\rangle$ a \textcolor{red}{power} $\langle$mask$\rangle$ $\langle$mask$\rangle$ greenhouse gases $\langle$mask$\rangle$   \\

\textbf{Pattern Noise (PN)}  &
\textcolor{red}{we} realized that \textcolor{red}{burnings} of fuels produces a large amount of greenhouse gases .     \\

\textbf{Backtranslation (BT)}  &
We realize that \textcolor{red}{burn} of \textcolor{red}{fuel produce} a \textcolor{red}{lot} of greenhouse gases .                    \\

\textbf{Round-translation (RT)}  &
We \textcolor{red}{recognize} that large amounts of greenhouse gases  \textcolor{red}{are generated by} combustion fuels .    \\

\bottomrule
\end{tabular}}

\caption{
Examples of pseudo source sentences generated by Direct Noise (DN), Pattern Noise (PN), Backtranslation (BT) and Round-translation (RT), respectively.
}
\label{tab:examples}

\end{table*}

\section{Introduction}\label{sec:intro}

Grammatical Error Correction (GEC) is a task that involves making \textit{locally substitutions} in text to correct all grammatical errors in a text~\cite{DBLP:journals/corr/abs-2305-10819, DBLP:journals/corr/abs-2307-09007, ye-etal-2023-system, DBLP:journals/corr/abs-2306-17447}.
GEC is often considered a monolingual machine translation (MT) task, which is typically tackled using sequence-to-sequence (Seq2Seq) architecture~\cite{bryant2022grammatical, DBLP:journals/csur/DongLGCLSY23}.
Numerous studies tend to improve the performance of Seq2Seq GEC models by increasing the amount of training data~\cite{DBLP:journals/corr/abs-2210-12692, DBLP:conf/acl/LiZLLLSWLCZ22, DBLP:conf/emnlp/LiMZLLHLLC022, 10095675}.

However, high-quality parallel data for GEC is not as widely available.
Despite the great success of Seq2Seq models, they are prone to overfitting and making predictions based on spurious patterns~\cite{tu-etal-2020-empirical}, owing to the vast gap between the number of model parameters and the limited high-quality data.
This data sparsity issue has motivated research into data augmentation in the field of GEC, particularly in the context of resource-heavy Seq2Seq approaches~\cite{rothe-etal-2021-simple,stahlberg-kumar-2021-synthetic}.
Thanks for the ease of constructing pseudo grammatical errors, recent studies focus on generating synthetic parallel data from clean monolingual corpora.
Various augmentation methods have been widely explored, whose major motivation comes from improving 
1) \textbf{distribution similarity} between pseudo and realistic data, including \textit{error patterns}~\cite{choe-etal-2019-neural} and \textit{back-translation}~\cite{sennrich-etal-2016-improving,xie-etal-2018-noising,stahlberg-kumar-2021-synthetic}; and 
2) \textbf{diversity} of pseudo data~\cite{koyama2021various}, such as \textit{noise injection}~\cite{grundkiewicz-etal-2019-neural,xu-etal-2019-erroneous} and \textit{round-trip translation}~\cite{zhou-etal-2020-improving-grammatical}.
However, prior works have primarily focused on showing the effectiveness of their proposed augmentation methods, without considering sample efficiency.
Training GEC models with excessive samples (e.g. over 100M~\cite{lichtarge-etal-2019-corpora,stahlberg-kumar-2021-synthetic}) for poor-scalable improvement is expensive and often unfeasible for most researchers.
Additionally, existing studies suffer from a lack of consistent experimental settings, making it intractable to systematically and fairly compare various data augmentation methods.
In this complex landscape, claims regarding distribution similarity and diversity remain unverified heuristics.
An uncomplicated approach to evaluate the effectiveness of an augmentation heuristic is to conduct experiments on all possible augmented datasets.
However, it is computationally expensive and even impractical when confronted with numerous augmentation heuristic options.

In this paper, to determine the extent to which distributional similarity and diversity of data augmentation can improve GEC models, we quantify both heuristics.
We use Affinity measure to evaluate the distribution similarity between pseudo and realistic grammatical errors, which is defined as the inverse of Kullback–Leibler (KL) divergence between them.
On the other hand, Diversity measure is used to assess the uncertainty of grammatical errors, defined as the entropy.
Next, we revisit four mainstream GEC data augmentation methods using our proposed measures.
Our findings illustrate that Affinity varies across these augmentation methods and is positively correlated with the performance of GEC models to some extent.
By altering the corruption rate of an augmentation method, we observe that the varying Diversity serves as a trade-off between Precision and Recall.

To overcome the challenge of limiting an augmentation strategy with high Affinity and appropriate Diversity, we propose MixEdit, a data augmentation approach that \textit{strategically} and \textit{dynamically} augments realistic data.
Unlike traditional approaches that rely on pseudo data generated from extra monolingual corpora, MixEdit regularizes over-parameterized GEC models using a limited amount of realistic data during fine-tuning.
MixEdit achieves this by replacing grammatical errors in the source sentence with other probable and label-preserving grammatical errors, avoiding undesired noise.
These augmented samples differ only in the form of grammatical errors, encouraging models to fully utilize the intrinsic information among diverse augmented samples, instead of learning spurious patterns during training.
We further apply Jensen-Shannon divergence consistency regularization to match the predictions between error patterns, and dynamically select candidate grammatical errors during fine-tuning.

We conduct experiments on two English GEC evaluation datasets: CoNLL-14~\cite{ng-etal-2014-conll} and BEA-19~\cite{bryant-etal-2019-bea}, and two Chinese GEC evaluation datasets: NLPCC-18~\cite{10.1007/978-3-319-99501-4_41} and MuCGEC~\cite{zhang-etal-2022-mucgec}.
Despite its simplicity, MixEdit consistently leads to significant performance gains compared to traditional methods.
By combining MixEdit with traditional methods, we achieve state-of-the-art results on BEA-2019, NLPCC-2018, and MuCGEC.

     


\section{Background}\label{sec:background}
To avoid confusion, we use the upper symbol $X$ to indicate a sentence, the bold lower symbol $\boldsymbol{x}$ to indicate a text segment.
In this paper, we focus the discussion on constructing grammatical errors for GEC data augmentation.
Generally, Seq2Seq-based GEC models learn the translation probability $P(Y\mid X;\boldsymbol{\theta})$, where $X$ denotes an ungrammatical source sentence and $Y$ represents a grammatical target sentence.
Given a parallel training dataset $\mathcal{D}$, the standard training objective is to minimize the empirical risk:

\begin{equation}
\mathcal L(\boldsymbol{\theta}) =\mathbb E_{(X,Y)\sim\mathcal{D}} [\mathcal{L}_{\operatorname{CE}}(X, Y; \boldsymbol{\theta})],
\label{eq:optimization}\end{equation}
where $\mathcal{L}_{\operatorname{CE}}$ is the cross entropy loss, $\mathcal{D}$ could be a realistic dataset $\mathcal{D}_r$ in a standard supervised learning setting, or a pseudo dataset $\mathcal{D}_p$ in typical GEC data augmentation settings, where the source sentences are usually generated from monolingual corpora~\cite{kiyono2020massive}.

Recent works have concentrated on improving the performance of GEC models by integrating various data augmentation techniques, which usually fall under the categories listed in Table~\ref{tab:examples}.

\paragraph{Direct Noise (DN).} DN injects noise into grammatically correct sentences in a rule-based manner~\cite{kiyono2020massive}.
The noise can take the form of 
1) masking, 
2) deletion, and 
3) insertion based on pre-defined probabilities.
DN is applicable to all languages since its rules are language-independent.
However, the generated errors are often not genuine and may even distort the original semantics of the sentences.

\paragraph{Pattern Noise (PN).} PN, on the other hand, involves injecting grammatical errors that are already present in the realistic GEC dataset into sentences~\cite{choe-etal-2019-neural}.
Specifically, this process entails first identifying error patterns in the GEC dataset using an automated error annotation tool such as ERRANT~\cite{bryant-etal-2017-automatic}, followed by applying a noising function that randomly substitutes text segments with grammatical errors.

\paragraph{Backtranslation (BT).} With the help of Seq2Seq models, BT can generate more genuine grammatical errors by learning the distribution of human-written grammatical errors~\cite{xie-etal-2018-noising}.
The noisy model is trained with the inverse of GEC parallel dataset, where ungrammatical sentence are treated as the target and grammatical ones as the source.
\citet{xie-etal-2018-noising} proposed several variants of BT, and showed that the variant \textbf{BT (Noisy)} achieved the best performance.
Therefore, we focus on this variant in this work.
When decoding the ungrammatical sentences, BT (Noisy) adds $r\beta_{\text{random}}$ to the score of each hypothesis in the beam for each time step, where $r$ is drawn uniformly from the interval $[0,1]$, and $\beta_{\text{random}}$ is a hyper-parameter that controls the degree of noise.

\paragraph{Round-translation (RT).} RT is an alternative method to generate pseudo data, which is based on the assumption that NMT systems may produce translation errors, resulting in noisy outputs via the bridge languages~\cite{lichtarge-etal-2019-corpora,zhou-etal-2020-improving-grammatical}.
The diverse outputs, however, may change the structure of the sentence due to the heterogeneity of different languages.

\paragraph{Training Settings.}
There are two primary training settings for incorporating $\mathcal{D}_p$ into the optimization of Equation~\eqref{eq:optimization}:
1) \textit{jointly optimizing} GEC models by concatenating the realistic dataset $\mathcal{D}_r$ and the pseudo datset $\mathcal{D}_p$, and 
2) \textit{pre-training} models on the pseudo dataset before fine-tuning on realistic datasets.
In a study conducted by~\citet{kiyono2020massive}, these two training settings were compared for two data augmentation methods (DN and BT), and it was found that pre-training was superior to joint optimization when large enough pseudo data was available.
Therefore, to avoid any adverse effects resulting from noisy augmented samples, we adopt the pre-training setting in our work.

\section{Method}\label{sec:method}

\subsection{Affinity and Diversity}\label{subsec:affinity_diversity}

\paragraph{Affinity.}
In-distribution corruption~\cite{choe-etal-2019-neural} has motivated the design of GEC data augmentation policies~\cite{grundkiewicz-etal-2019-neural,stahlberg-kumar-2021-synthetic,yasunaga-etal-2021-lm}, based on the idea that pseudo data with less distribution shift should improve performance on specific evaluation datasets.
Inspired by this focus, we propose Affinity, which is used to qualify how augmentation shifts data with respect to the error pattern.
We define Affinity of a data augmentation method as the inverse of Kullback–Leibler (KL) divergence between pseudo and realistic grammatical errors, which can be computed as follow:

\begin{equation}\begin{aligned}
\frac{1}{\operatorname{Affinity}(\mathcal D_s,\mathcal{D}_r)}= \frac{1}{2}\operatorname{KL}(P_p(\boldsymbol{x}, \boldsymbol{y}) \parallel P_r(\boldsymbol{x}, \boldsymbol{y})) \\
\quad + \frac{1}{2} \operatorname{KL}(P_r(\boldsymbol{x}, \boldsymbol{y}) \parallel P_p(\boldsymbol{x}, \boldsymbol{y})),
\end{aligned}\end{equation}
where $\mathcal{D}_p$ and $\mathcal{D}_r$ refer to the pseudo and realistic datasets, respectively.
The pair of text corrections, denoted by $\boldsymbol{x}$ and $\boldsymbol{y}$, can be extracted using an automated error annotation toolkit such as ERRANT~\cite{bryant-etal-2017-automatic}.
$P_p$ and $P_r$ denote the probabilities of pseudo and realistic grammatical errors, and $\operatorname{KL}$ represents KL divergence used to calculate the distance between two distributions.
To prevent $\operatorname{KL}$ approaching infinity, we limit the support set of the calculation.
The first term of the above equation is computed as follow:

\begin{equation}\begin{aligned}
&\operatorname{KL}(P_p(\boldsymbol{x}, \boldsymbol{y}) \parallel P_r(\boldsymbol{x}, \boldsymbol{y})) = \\
&\quad\quad\quad\sum_{(\boldsymbol{x},\boldsymbol{y})\sim\mathcal{D}_r} P_p(\boldsymbol{x},\boldsymbol{y})\log\frac{P_p(\boldsymbol{x}, \boldsymbol{y})}{P_r(\boldsymbol{x}, \boldsymbol{y})}.
\end{aligned}\end{equation}

With this definition, Affinity of a small value suggests the pseudo grammatical errors are more likely to be distributed within the realistic dataset.

\begin{figure*}[ht!]
\centering
\includegraphics[scale=0.34]{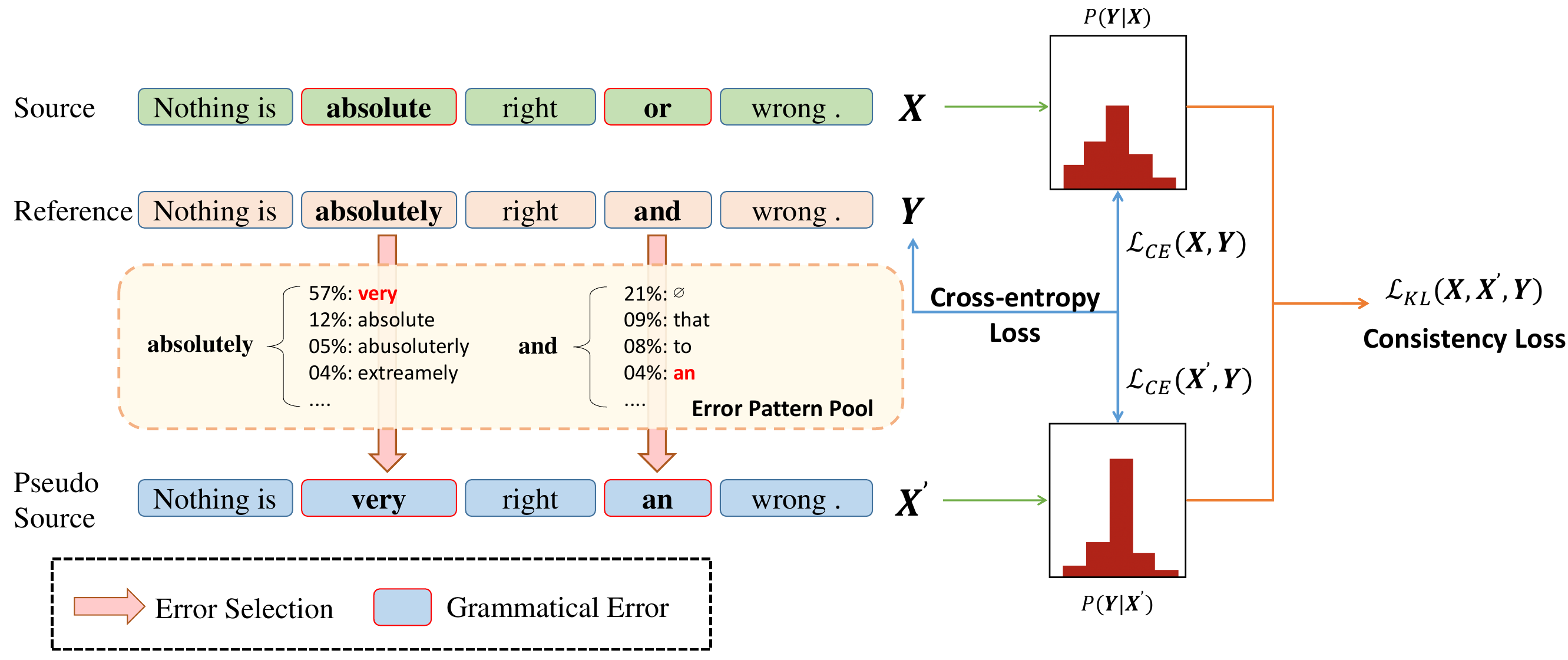}
\caption{
Overview of our approach MixEdit. 
MixEdit 
1) first extracts the error patterns from GEC realistic datasets and builds \textbf{Error Pattern Pool}, 
2) replaces grammatical errors with alternative candidates from the Error Pattern Pool, and then 
3) computes the cross-entropy loss $\mathcal{L}_{\operatorname{CE}}$ and the consistency loss $\mathcal{L}_{\operatorname{KL}}$.
}
\label{fig:mixedit}
\end{figure*}
\paragraph{Diversity.}
\citet{koyama2021various} demonstrated the importance of diversity in pseudo grammatical errors for improving performance.
Based on the observation, we introduce another perspective to evaluate GEC data augmentation policies, which we dub Diversity, defined as the entropy of pseudo grammatical errors:

\begin{equation}\begin{aligned}
\operatorname{Diversity}=-\sum_{(\boldsymbol{x},\boldsymbol{y})}P(\boldsymbol{x},\boldsymbol{y})\log P(\boldsymbol{x},\boldsymbol{y}),
\end{aligned}\end{equation}
where $P(\cdot)$ can take on the values of $P_p(\cdot)$ or $P_r(\cdot)$.
Like Affinity, this definition of Diversity has the advantage of capturing task-specific and interpretable elements, i.e., the error patterns of GEC datasets.
Furthermore, these measures are off-the-shelf and require little computational cost.

It should be kindly noted that the performance of GEC models is not purely a function of training data, as training dynamics and implicit biases in the model can also impact final performance.
Both of these measures are introduced to provide a new perspective on characterizing and understanding GEC data augmentation.

\subsection{MixEdit}\label{subsec:mixedit}


MixEdit aims to strategically and dynamically construct pseudo data with high Affinity and appropriate Diversity, thus achieving the best of both worlds.
Specifically, we first extract error patterns from a GEC dataset $\mathcal{D}$.
The following corruption probability can be derived from the error patterns using Bayes' rule:

\begin{equation}
    P(\boldsymbol{x}\mid\boldsymbol{y})=\frac{P(\boldsymbol{x}, \boldsymbol{y})}{P(\boldsymbol{y})}.
\end{equation}

The corruption probability describes the correction feature of GEC dataset.
For each parallel sample $(\mathbf{x}, \mathbf{y})$ the correction can be denoted by an edit sequence $\mathbf{e}=\{\boldsymbol{e}_1,\boldsymbol{e}_2,\cdots,\boldsymbol{e}_m\}$, where each edit $\boldsymbol{e}_i$ consists of source tokens $\boldsymbol{x}$ and target tokens $\boldsymbol{y}$.
As shown in Figure~\ref{fig:mixedit}, MixEdit dynamically replace the original source tokens $\boldsymbol{x}$ with other candidates $\boldsymbol{x}'$ from a pool of error patterns during training.
Since all the candidates share the same target tokens $\boldsymbol{y}$, this transformation is typically label-preserving, preventing undesired noise.
These label-preserving perturbations provides various literal forms of the same grammatical errors, enabling correct predictions based on more complex contexts instead of spurious patterns.

Inspired by~\citet{shen2020simple,chen-etal-2021-hiddencut}, we incorporate consistency regularization in our model to encourage stable and similar predictions across realistic and augmented samples.
The training objective is written as:~\footnote{\citet{shen2020simple} considered multiple augmented samples for a realistic sample, but we found a single augmented sample works just as well.
Additionally, increasing the number of augmented samples significantly escalates the training cost.
}

\begin{equation}\begin{aligned}
\mathcal{L} = & \mathcal{L}_{\operatorname{CE}}(X, Y) + \alpha \mathcal{L}_{\operatorname{CE}}(X', Y) \\
& + \beta \mathcal{L}_{\operatorname{KL}}(X, X', Y),
\end{aligned}\label{eq:loss}\end{equation}
where the augmented sample $X'$ is generated dynamically during training, with weights $\alpha$ and $\beta$ used to balance the contribution of learning from the original data and the augmented data.
$\mathcal{L}_{\operatorname{CE}}$ denotes the cross-entropy loss, and $\mathcal{L}_{\operatorname{KL}}$ is KL-divergence as follow:

\begin{equation}
\mathcal{L}_{\operatorname{KL}}(X, X', Y) = \operatorname{KL}\left[ P(Y \mid X') \parallel P_{avg} \right],
\end{equation}
where $P_{avg}$ represents the average prediction probability across realistic and augmented samples.

PN is the most similar traditional GEC data augmentation method to MixEdit.
However, there are several distinctions between these two methods: 
1) MixEdit provides different views for the same grammatical errors, while PN randomly constructs pseudo errors.
In our preliminary experiments~\footnote{
We leave more detailed analysis to Section~\ref{subsec:analysis}.
}, we found it matters to determine the positions for constructing grammatical errors.
2) PN is typically used to generate new pseudo parallel data from monolingual corpora~\cite{grundkiewicz-junczys-dowmunt-2019-minimally,white-rozovskaya-2020-comparative}.
Conversely, MixEdit is proposed to augment high-quality realistic data and can be combined with other traditional data augmentation methods introduced in Section~\ref{sec:background} due to their orthogonality.

\section{Experiments on English GEC}\label{sec:experiments_english}

\subsection{Experimental Settings}

\paragraph{Datasets and evaluation.}
We decompose the training of the baseline model into three stages following~\citet{zhang-etal-2022-syngec}.
We train the model on 
1) the CLang8 dataset~\cite{rothe-etal-2021-simple},
2) the FCE dataset~\cite{yannakoudakis-etal-2011-new}, the NUCLE dataset~\cite{dahlmeier-etal-2013-building} and the W\&I+LOCNESS train-set~\cite{bryant-etal-2019-bea}.
3) We finally fine-tune the model on high-quality W\&I-LOCNESS.
As for the traditional data augmentation methods using extra corpora (i.e., DN, PN, BT and RT), we construct a pseudo dataset using the seed corpus Gigaword~\footnote{
\url{https://catalog.ldc.upenn.edu/LDC2011T07}
}, which has been proven to be the best among three seed corpora by~\citet{kiyono2020massive}.
In total, we generate 8M pseudo data with the same target sentences for four data augmentation methods, which is used for the pre-training of GEC models.

For evaluation, we reports the results on the CoNLL-14 test set~\cite{ng-etal-2014-conll} evaluated by M$^2$ scorer~\cite{dahlmeier-ng-2012-better} and the BEA-19 test set~\cite{bryant-etal-2019-bea} evaluated by ERRANT.
The results are averaged over three runs with different random seeds, and the BEA-19 dev set serves as the validation set.
We provide statistics of all the involved datasets in Table~\ref{tab:dataset_eng}.

\paragraph{GEC backbone model.}
We adopt Transformer-based BART-Large~\cite{lewis-etal-2020-bart} as our backbone model, which has been shown as a strong baseline for GEC~\cite{katsumata-komachi-2020-stronger,zhang-etal-2022-syngec}.
We acquire subwords by byte-pair-encoding (BPE)~\cite{sennrich-etal-2016-neural} algorithm.
We apply the Dropout-Src mechanism~\cite{junczys-dowmunt-etal-2018-approaching} to source-side word embeddings following~\citet{zhang-etal-2022-syngec}.
All experiments are conducted using the \texttt{Fairseq}~\cite{ott-etal-2019-fairseq} public toolkit.
Most of the hyperparameter settings are identical to~\citet{zhang-etal-2022-syngec}, which are provided in Appendix~\ref{appendix:hyper_parameters}.

\paragraph{Data Augmentation.}
We examine and analyze four mainstream GEC data augmentation methods discussed in Section~\ref{sec:background}, as well as our proposed MixEdit in Section~\ref{subsec:mixedit}.
We introduce an extra pre-training stage that utilizes the pseudo datasets generated by each data augmentation method.
Further details regarding the experimental settings for augmentation methods can be found in Appendix~\ref{appendix:details_augmentation}.

\begin{table}[tp!]
\scalebox{0.86}{
\begin{tabular}{lccc}
\toprule
\textbf{Dataset} & \textbf{\#Sentences} & \textbf{Usage}    \\ 
\hline

\textbf{Gigaword}         & 8,000,000       & Pre-training        \\
\textbf{CLang8}           & 2,372,119       & Fine-tuning I       \\ 
\textbf{FCE}              & 34,490          & Fine-tuning II     \\
\textbf{NUCLE}            & 57,151          & Fine-tuning II     \\
\textbf{W\&I+LOCNESS}     & 34,308          & Fine-tuning II\&III \\
\hline

\textbf{BEA-2019-\textit{Dev}}      & 4,384     & Validation  \\ 
\textbf{BEA-2019-\textit{Test}}     & 4,477     & Testing     \\
\textbf{CoNLL-2014-\textit{Test}}   & 1,312     & Testing     \\
 
\bottomrule
\end{tabular}}

\caption{
Statistics of English GEC datasets.
Gigaword is only available for DN, PN, BT and RT.
}
\label{tab:dataset_eng}

\end{table}
\begin{table*}[tp!]
\centering
\scalebox{0.75}{
\begin{tabular}{clcccccccc}
\toprule

&  &  \textbf{Extra}  &  \multicolumn{1}{c}{\textbf{Transformer}}  &  \multicolumn{3}{c}{\textbf{CoNLL-14-\textit{test}}}  &  \multicolumn{3}{c}{\textbf{BEA-19-\textit{test}}}     \\

& \textbf{System}  & \textbf{Data Size} & \textbf{Layer, Hidden, FFN}  &  \textbf{P}  &  \textbf{R} & $\mbox{\textbf{F}}_{0.5}$ & \textbf{P} & \textbf{R} &$\mbox{\textbf{F}}_{0.5}$ \\

\hline

\multirow{3}{*}{\textbf{w/o PLM}}

& \citet{kiyono-etal-2019-empirical}$^{\circ}$ 
&  70M   &  12+12,1024,4096         
&  67.9  &  44.1  &  61.3    &  65.5  &  59.4  &  64.2          \\

& \citet{lichtarge-etal-2020-data}$^{\vartriangle\blacktriangle}$
&  340M  &  12+12,1024,4096
&  69.4  &  43.9  &  62.1    &  67.6  &  62.5  &  66.5          \\

& \citet{stahlberg-kumar-2021-synthetic}$^{\vartriangle\blacktriangle\square}$  &  540M  &  12+12,1024,4096
&  72.8  &  49.5  & \textbf{66.6}    &  72.1  &  64.4  & \textbf{70.4}          \\

\hline \hline

\multirow{17}{*}{\textbf{w/ PLM}}

& \citet{kaneko-etal-2020-encoder}$^{\circ}$
&  70M   &   12+12,1024,4096
&  69.2  &  45.6  &  62.6    &  67.1  &  60.1  &  65.6          \\

& \citet{katsumata-komachi-2020-stronger}
&  -     &   12+12,1024,4096
&  69.3  &  45.0  &  62.6    &  68.3  &  57.1  &  65.6          \\

& \citet{omelianchuk-etal-2020-gector}$^{\lozenge}$
&  9M    &   12+0,768,3072
&  77.5  &  40.1  &  65.3    &  79.2  &  53.9  &  72.4          \\

& \citet{rothe-etal-2021-simple}$^{\heartsuit}$
&  2.4M  &  12+12,1024,4096
&  -     &  -     &  66.1    &  -     &  -     &  72.1          \\

& \citet{sun-etal-2021-instantaneous}$^{\bigstar}$
&  300M  &  12+2,1024,4096
&  71.0  &  52.8  &  66.4    & -      & -      &  72.9          \\

& \textbf{BART Baseline}$^{\heartsuit}$~\cite{zhang-etal-2022-syngec}
&  2.4M  &  12+12,1024,4096 
&  73.6  & 	48.6  &  66.7    &  74.0  &  64.9  &  72.0          \\

& \textbf{SynGEC}$^{\heartsuit}$~\cite{zhang-etal-2022-syngec}
&  2.4M  &  12+12,1024,4096 
&  74.7  & 	49.0  &  \textbf{67.6}    &  75.1  &  65.5  &  \textbf{72.9}          \\

\cline{2-10} 

& \textbf{Our BART Baseline}$^{\heartsuit}$
&  2.4M  &  12+12,1024,4096
&  73.8  &  47.6  &  66.5    &  74.4  &  63.7  &  72.0          \\

& \hspace{0.3cm} \textbf{+}\textbf{DN}$^{\clubsuit}$$^{\heartsuit}$
&  10.4M  &  12+12,1024,4096
&  70.3  &  50.3  &  65.1    &  72.6  &  64.6  &  70.9          \\

& \hspace{0.3cm} \textbf{+}\textbf{PN}$^{\clubsuit}$$^{\heartsuit}$
&  10.4M  &  12+12,1024,4096
&  73.3  &  51.1  &  \textbf{67.4}    &  75.0  &  65.1  &  72.8          \\

& \hspace{0.3cm} \textbf{+}\textbf{BT}$^{\clubsuit}$$^{\heartsuit}$
&  10.4M  &  12+12,1024,4096
&  72.7  &  51.3  &  67.1    &  75.1  &  65.9  &  \textbf{73.0}          \\

& \hspace{0.3cm} \textbf{+}\textbf{RT}$^{\clubsuit}$$^{\heartsuit}$
&  10.4M  &  12+12,1024,4096
&  73.0  &  48.7  &  66.4    &  75.1  &  64.0  &  72.5          \\

\cline{2-10}

& \textbf{MixEdit}$^{\heartsuit}$
&  2.4M  &  12+12,1024,4096
&  75.6  &  46.8  &  67.3$^\uparrow$    &  76.4  &  62.7  &  73.2$^\uparrow$          \\
& \hspace{0.3cm} \textbf{+}\textbf{DN}$^{\clubsuit}$$^{\heartsuit}$
&  10.4M  &  12+12,1024,4096
&  72.6  &  48.3  &  66.0$^\uparrow$    &  74.4  &  63.0  &  71.8$^\uparrow$          \\

& \hspace{0.3cm} \textbf{+}\textbf{PN}$^{\clubsuit}$$^{\heartsuit}$
&  10.4M  &  12+12,1024,4096
&  75.1  &  48.1  &  \textbf{67.5}$^\uparrow$    &  75.3  &  64.7  &  72.9$^\uparrow$          \\

& \hspace{0.3cm} \textbf{+}\textbf{BT}$^{\clubsuit}$$^{\heartsuit}$
&  10.4M  &  12+12,1024,4096
&  74.4  &  48.4  &  67.2$^\uparrow$    &  76.4  &  63.7  &  \textbf{73.4}$^\uparrow$          \\

& \hspace{0.3cm} \textbf{+}\textbf{RT}$^{\clubsuit}$$^{\heartsuit}$
&  10.4M  &  12+12,1024,4096
&  75.2  &  47.2  &  67.2$^\uparrow$    &  75.9  &  63.3  &  72.9$^\uparrow$          \\

\bottomrule
\end{tabular}}

\caption{
The results of baselines and GEC data augmentation methods for \textbf{single-model}.
\textbf{Layer}, \textbf{Hidden} and \textbf{FFN} denote the depth, hidden size and feed-forward network size of Transformer.
``Our BART Baseline'' is re-implemented from the open-source ``BART Baseline'', both of which are under the same experimental setting, making them fairly comparable.
$^\uparrow$means the performance of "\textbf{w/ MixEdit}" is better than that of its "\textbf{w/o MixEdit}" counterpart.
Besides the public human-annotated training data, private and/or pseudo data sources are also widely used in GEC systems, including: 
$^{\circ}$BT pseudo data from Gigaword (70M sentences),
$^{\vartriangle}$Wikipedia revision histories (170M),
$^{\blacktriangle}$RT pseudo data from Wikipedia (170M),
$^{\square}$BT pseudo data from Colossal Clean Crawled Corpus (200M),
$^{\lozenge}$DN psuedo data from one-billion-word (9M),
$^{\bigstar}$BT and DN pseudo data (300M),
$^{\heartsuit}$cleaned version of Lang8 (2.4M),
$^{\clubsuit}$our psuedo data from Gigaword (8M).
}
\label{tab:main:results}
\end{table*}

\subsection{Results of GEC Data Augmentation}\label{results_eng}
Table~\ref{tab:main:results} showcases the results of each data augmentation method.
Based on the baseline model, PN achieves the highest F$_{0.5}$ score on CoNLL-14, while BT obtains the highest F$_{0.5}$ score on BEA-19.
In contrast, DN produces lower F$_{0.5}$ score than the baseline in both test sets, which contradicts the findings of~\citet{kiyono-etal-2019-empirical} that DN can benefit the GEC performance of scratch transformer models.
We attribute the phenomenon to the fact that the pre-training task of BART is similar to DN, and it is unnecessary and possibly harmful to pre-train BART again with task-independent noise.
Finally, RT performs slightly better on BEA-19 but slightly worse on CoNLL-14.

The bottom group lists the results of MixEdit and its combination with each traditional GEC data augmentation method.
Without extra monolingual corpora, our MixEdit achieves 67.3/73.2 F$_{0.5}$ scores, which are on par with or marginally superior to traditional methods using the same backbone model.
Notably, MixEdit also complements these traditional methods.
Each augmentation method based on MixEdit produces higher F$_{0.5}$ scores than its counterpart, with PN and BT \textbf{w/ MixEdit} achieving the highest F$_{0.5}$ scores on CoNLL-14 (67.5) and BEA-19 (73.4), respectively.

\subsection{Analysis}\label{subsec:analysis}

The mechanism through which these corruptions work remains unclear, although the effectiveness of GEC data augmentation is well-established.
In this section, we investigate the relationship between data augmentation and performance through the lens of quantified Affinity and Diversity measures, seeking to gain insight into the mechanisms underlying GEC data augmentation.

To ensure a fair comparison, we avoid introducing extra monolingual corpora that may influence result due to nuisance variables such as text domain.
Instead, we apply data augmentation to GEC realistic datasets comprising FCE, NUCLE and W\&I+LOCNESS (collectively referred to as BEA-train).
Specifically, we retain the target sentences from BEA-train, and generate pseudo source sentences by enforcing data augmentation strategies, resulting in a set of pseudo datasets with identical targets but different sources.
It is worth noting that we remove the dynamic training and the regularization of MixEdit to facilitate a fair comparison with other data augmentation methods.
We respectively train GEC models on these pseudo datasets and report the results in Table~\ref{tab:affinity_diversity_bea}.

\paragraph{Affinity is positively correlated with the performance.}
At a high level, Affinity of MixEdit, PN and BT, which achieve the highest F$_{0.5}$ scores, is much higher than that of DN and RT.
We compute the Pearson's correlation coefficient for the data augmentation methods involved: DN, PN, BT, RT and MixEdit.
For PN and BT, we choose the hyper-parameter configurations that yield the highest F0.5 scores.
The Pearson's correlation coefficient between F$_{0.5}$ and Affinity is 0.9485.
The results indicate a strong correlation between Affinity and F$_{0.5}$ on BEA-train.
It should be noted that, despite having lower Affinity, BT achieves a higher F$_{0.5}$ score than PN.
We attribute this to the advantage of BT in learning the distribution of grammatical errors using Seq2Seq models instead of adding errors crudely.
MixEdit also skillfully avoids the drawbacks of PN by strategically applying label-preserving perturbations, resulting in an approximate F$_{0.5}$ score with the baseline.

\begin{table}[tp!]
\centering
\scalebox{0.7}{
\begin{tabular}{lllccc}
\toprule

\textbf{Method}  &  \textbf{Affinity}$^\uparrow$  &  \textbf{Diversity}  &
\textbf{P}  &  \bf{R} & $\mbox{\textbf{F}}_{0.5}$           \\
\hline

\textbf{Baseline}   &  -     &  8.78   &  \bf{56.03}  &  37.60  &  \bf{51.03}     \\
\textbf{DN}  &  0.41  &  10.70  &  20.49  &  12.94  &  18.35     \\
\textbf{RT}  &  0.75  &  10.54  &  20.59  &  38.41  &  22.70     \\

\textbf{PN}   \\
\hspace{0.3cm} \textbf{Round=1}  &  1.91  &  7.46  &  35.46  &  16.35  &  28.74     \\
\hspace{0.3cm} \textbf{Round=2}  &  \bf{1.93}  &  7.56  &  \bf{39.43}  &  23.71  &  34.82     \\
\hspace{0.3cm} \textbf{Round=4}  &  1.90  &  7.70  &  39.41  &  29.41  &  36.90     \\
\hspace{0.3cm} \textbf{Round=8}  &  1.77  &  7.92  &  42.14  &  37.51  &  \textbf{41.13}     \\
\hspace{0.3cm} \textbf{Round=16} &  1.59  &  \bf{8.24}  &  33.83  &  \bf{39.24}  &  34.79     \\

\textbf{BT}   \\
\hspace{0.3cm} \textbf{$\beta_{\operatorname{random}}$=0}
&  0.65  &  6.96  &  22.20  &  7.36   &  15.82      \\
\hspace{0.3cm} \textbf{$\beta_{\operatorname{random}}$=3}
&  1.47  &  7.67  &  44.11  &  32.46  &  41.16      \\
\hspace{0.3cm} \textbf{$\beta_{\operatorname{random}}$=6}
&  \bf{1.57}  &  8.22  &  \bf{49.29}  &  42.99  &  \bf{47.89}      \\
\hspace{0.3cm} \textbf{$\beta_{\operatorname{random}}$=9}
&  1.53  &  8.59  &  44.69  &  47.76  &  45.27      \\
\hspace{0.3cm} \textbf{$\beta_{\operatorname{random}}$=12}
&  1.45  &  \bf{8.84}  &  41.43  &  \bf{50.64}  &  43.00      \\

\textbf{MixEdit} (Ours)   &  \textbf{2.33}  &  8.52   &  \bf{57.72}  &  33.24  &  \textbf{50.31}      \\

\bottomrule
\end{tabular}}

\caption{
Affinity and Diversity of data augmentation methods.
The baseline model is trained using realistic BEA-train dataset.
}
\label{tab:affinity_diversity_bea}

\end{table}

\paragraph{Diversity is responsible for the trade-off between Precision and Recall.}
We qualitatively investigate the effect of Diversity on GEC performance within a fixed method.
The trade-off of Diversity is apparent when adjusting certain hyper-parameters responsible for Diversity (Round for PN and $\beta$ for BT).
As Diversity increases, Precision and F$_{0.5}$ initially increase and then decrease, reaching their peak at an appropriate setting.
Meanwhile, Recall continues to increase.
For example, PN and BT reach their peak F$_{0.5}$ score at \textit{Round=8} and \textit{$\beta_{\operatorname{random}}$=6}, respectively, falling at an intermediate value of Diversity.

Therefore, we argue that an excellent data augmentation technique should have high Affinity and appropriate Diversity, which motivated our proposed MixEdit.
The high Affinity of MixEdit stems from the fact that the distribution of pseudo grammatical errors it generates is the same as that of the ground truth.
MixEdit only replaces the original grammatical errors with augmented ones, maintaining the same error density in the augmented dataset.
As a result, the diversity of the augmented dataset closely resembles that of the original dataset.

To further explore the relationship between Affinity/Diversity and performance across different datasets, we also conduct additional experiments on English CLang8 and Chinese HSK.
The relationship between Affinity/Diversity and performance is similar to the results on BEA-train, which are provided in Appendix~\ref{app:extra_datasets}.
Additionally, we provide further analysis on the complementary effectiveness of pseudo data in Appendix~\ref{app:complement_psuedo_data}.

\begin{table}[tp!]
\centering
\scalebox{0.66}{
\begin{tabular}{lcc}
\toprule

&  \textbf{CoNLL-14-\textit{test}}  &  \textbf{BEA-19-\textit{test}} \\ 
& \textbf{P}/\textbf{R}/\textbf{$\mbox{\textbf{F}}_{0.5}$} & \textbf{P}/\textbf{R}/\textbf{$\mbox{\textbf{F}}_{0.5}$} \\

\hline

\textbf{MixEdit}  
&  76.81/45.00/67.30  &  76.37/62.71/\textbf{73.18}      \\

\hspace{0.3cm}\textbf{w/o Error Pattern}   
&  73.08/48.46/66.35  &  75.22/63.08/72.44      \\

\hspace{0.3cm}\textbf{w/o Consistency Loss}
&  74.73/47.80/67.16  &  76.04/63.39/73.12      \\

\hspace{0.3cm}\textbf{w/o Dynamic Generation}  
&  75.41/46.74/67.17  &  75.73/62.35/72.61      \\

\hspace{0.3cm}\textbf{w Pattern Noise}
&  74.70/49.73/\textbf{67.88}  &  74.78/64.67/72.51      \\

\bottomrule
\end{tabular}}
\caption{
Ablation results of MixEdit.
}
\label{tab:ablation_module}
\end{table}
\begin{table}[tp!]
\centering
\scalebox{0.80}{
\begin{tabular}{lccc}
\toprule

$\alpha$  & \bf{P}  &  \bf{R}  &  \bf{$\mbox{\textbf{F}}_{0.5}$}  \\

\hline

\bf{0.5}  &  55.96  &  41.35  &  52.27  \\
\bf{0.8}  &  55.65  &  41.35  &  52.05  \\
\bf{1.0}  &  57.96  &  39.00  &  \bf{52.83}  \\
\bf{1.2}  &  56.75  &  40.93  &  52.68  \\
\bf{2.0}  &  55.88  &  41.30  &  52.20  \\

\bottomrule
\end{tabular}}
\caption{
Results of various $\alpha$ on BEA-train.
}
\label{tab:ablation_alpha}
\end{table}
\begin{table}[tp!]
\centering
\scalebox{0.80}{
\begin{tabular}{lccc}
\toprule

$\beta$  & \bf{P}  &  \bf{R}  &  \bf{$\mbox{\textbf{F}}_{0.5}$}  \\

\hline

\bf{0.5}  &  57.56  &  38.51  &  52.38  \\
\bf{1.0}  &  57.96  &  39.00  &  \bf{52.83}  \\
\bf{2.0}  &  56.69  &  40.59  &  52.52  \\

\bottomrule
\end{tabular}}
\caption{
Results of various $\beta$ on BEA-train.
}
\label{tab:ablation_beta}
\end{table}
\begin{table*}[tp!]
\centering
\scalebox{0.8}{
\begin{tabular}{lccccccc}
\toprule

&  \textbf{Extra}  
&  \multicolumn{3}{c}{\textbf{NLPCC-18-\textit{test}}}  
&  \multicolumn{3}{c}{\textbf{MuCGEC-\textit{test}}}            \\

\textbf{System}  &  \textbf{Data Size}  
&  \textbf{P}  &  \textbf{R} &  $\mbox{\textbf{F}}_{0.5}$ 
&  \textbf{P}  &  \textbf{R} &  $\mbox{\textbf{F}}_{0.5}$       \\

\hline

\textbf{\citet{zhang-etal-2022-mucgec}}  &  -  &
42.88  &  30.19  &  39.55  &  43.81  &  28.56  &  39.58         \\

\textbf{SynGEC}~\cite{zhang-etal-2022-syngec}  &  -  &
49.96  &  33.04  &  45.32  &  54.69  &  29.10  &  46.51         \\

\hline

\textbf{Our Baseline}  &  -  &
49.81  &  31.57  &  44.65    &  54.24  &  29.67  &  46.53         \\

\hspace{0.3cm} \textbf{+}\textbf{DN}  &  8M  &
49.57  &  31.80  &  44.59    &  54.93  &  29.61  &  46.91          \\

\hspace{0.3cm} \textbf{+}\textbf{PN}  &  8M  &
50.15  &  35.27  &  \textbf{46.25}    &  55.83  &  30.15  &  \textbf{47.71}          \\

\hspace{0.3cm} \textbf{+}\textbf{BT}  &  8M  &
47.64  &  36.43  &  44.88    &  54.82  &  30.27  &  47.17          \\

\hspace{0.3cm} \textbf{+}\textbf{RT}  &  8M  &
51.06  &  30.74  &  45.09    &  56.96  &  27.41  &  46.86         \\

\hline

\textbf{MixEdit}  &  -  &
49.58  &  32.93  &  45.03$^\uparrow$    &  55.25  &  29.30  &  46.94$^\uparrow$          \\

\hspace{0.3cm} \textbf{+}\textbf{DN}  &  8M  &
50.46  &  30.55  &  44.64$^\uparrow$    &  56.48  &  28.12  &  47.00$^\uparrow$          \\

\hspace{0.3cm} \textbf{+}\textbf{PN}  &  8M  &
52.26  &  33.37  &  \textbf{46.94}$^\uparrow$    &  56.99  &  29.73  &  \textbf{48.16}$^\uparrow$          \\

\hspace{0.3cm} \textbf{+}\textbf{BT}  &  8M  &
48.99  &  35.45  &  45.52$^\uparrow$    &  54.72  &  31.76  &  47.81$^\uparrow$          \\

\hspace{0.3cm} \textbf{+}\textbf{RT}  &  8M  &
51.19  &  34.94  &  45.68$^\uparrow$    &  55.04  &  30.53  &  47.43$^\uparrow$          \\

\bottomrule

\end{tabular}}

\caption{
\textbf{Single-model} results on Chinese datasets.
All models are initialized with pre-trained Chinese BART weights.
$^\uparrow$ means the performance of "\textbf{w/ MixEdit}" is better than its "\textbf{w/o MixEdit}" counterpart.
}
\label{tab:chinese:result}

\end{table*}
\begin{table}[tp!]
\centering
\scalebox{0.8}{
\begin{tabular}{lcc}
\toprule

\textbf{Dataset}  &  \textbf{\#Sentences}  &  \textbf{Usage}    \\ 
\hline

\textbf{News2016zh}     &  8,000,000  & Pre-training        \\
\textbf{Lang8}          &  1,220,906  & Fine-tuning         \\ 
\textbf{HSK}            &  156,870    & Fine-tuning         \\ 

\hline

\textbf{MuCGEC-\textit{dev}}        & 1,125  & Validation      \\ 
\textbf{MuCGEC-\textit{test}}       & 5,938  & Testing         \\ 
\textbf{NLPCC-18-\textit{test}}     & 2,000  & Testing         \\ 

\bottomrule
\end{tabular}}

\caption{Statistics of Chinese GEC datasets.}
\label{tab:chinese:dataset}

\end{table}

\subsection{Ablation Study}

\paragraph{Decomposition of MixEdit.}
We explore the effectiveness of each component of our proposed MixEdit by conducting ablation studies shown in Table~\ref{tab:ablation_module}.
Specifically, for the "w/o Error Pattern" variant, we randomly mask tokens in the sentence instead of sampling grammatical errors from the Error Pattern Pool.
For "w/o Consistency Loss", we remove the consistency loss.
For "w/o Dynamic Generation", we always generate fixed pseudo data for fine-tuning GEC models.
Additionally, we attempt to incorporate PN into generating dynamic pseudo data in the fine-tuning stage.
The results demonstrate that task-specific information of error pattern is important to constructing high-quality pseudo data.
Dynamic Generating is another critical factor for the success of MixEdit, since it can improve the diversity of pseudo data without the loss of Affinity.
Surprisingly, adding pattern noise improves F$_{0.5}$ on CoNLL-14-\textit{test}, with a performance reduction on BEA-19-\textit{test}.
Given that CoNLL-14 is centered around essays written by language learners, our speculation is that it includes a greater number of stereotypical grammatical errors.
As a result, incorporating pseudo data with a higher corruption ratio can improve performance on CoNLL-14.
We investigate how varying PN corruption ratios affect the performance of the GEC model in Appendix~\ref{app:pn+mixedit}.

\paragraph{Sensitivity to hyperparameters.}
We investigate the sensitivity to the choice of the hyperparameters $\alpha$ and $\beta$ introduced in Equation~\ref{eq:loss}.
We explore the optimal values of them on BEA-train.
The results of various $\alpha$ and $\beta$ are reported in Table~\ref{tab:ablation_alpha} and Table~\ref{tab:ablation_beta}, respectively, where the optimal hyperparameter setting is $\alpha=1.0$ and $\beta=1.0$.

\section{Experiments on Chinese GEC}\label{sec:experiments_chinese}
\subsection{Experimental Settings}

\paragraph{Datasets and evaluation.}
We adopt the seed corpus \textit{news2016zh}~\footnote{
\url{https://github.com/brightmart/nlp\_chinese\_corpus}
} to generate 8M pseudo data for each traditional data augmentation method, similar to our English experiments.
Following~\cite{zhang-etal-2022-mucgec}, we fine-tune GEC models on the Chinese Lang8~\cite{10.1007/978-3-319-99501-4_41} and HSK~\cite{zhang2009features} datasets.
We reports the results on NLPCC-2018-$\textit{test}$~\cite{10.1007/978-3-319-99501-4_41} evaluated by M$^2$ scorer, and MuCGCE-$\textit{test}$~\cite{zhang-etal-2022-mucgec} evaluated by ChERRANT.
Table~\ref{tab:chinese:dataset} provides statistics for the aforementioned datasets.

\paragraph{GEC backbone model.}
We employ Chinese BART~\cite{shao2021cpt} as our backbone model~\footnote{
\url{https://huggingface.co/fnlp/bart-large-chinese}
}.
We retained the original vocabulary because the updated version of Chinese BART has already incorporated a larger vocabulary, and therefore no modifications were necessary.

\subsection{Results}\label{results_zho}

Table~\ref{tab:chinese:result} presents our results.
With the exception of PN, all data augmentation methods improve the F$_{0.5}$ score, demonstrating that low-affinity data augmentation methods can negatively impact the performance of pre-trained models like BART.
Moreover, incorporating MixEdit with traditional methods can further increase the F$_{0.5}$ scores on both evaluation datasets.
These findings suggest that MixEdit is a general technique that can be effectively employed in various languages and combined with other data augmentation methods.

\section{Related Works}\label{sec:related_works}

\paragraph{Data Augmentation.}
Data augmentation encompasses methods of increasing training data diversity without requiring additional data collection.
It has become a staple component in many downstream NLP tasks.
Most researches apply a range of perturbation techniques to increase training scale, with the aim of reducing overfitting and improving the generalization of models~\cite{feng-etal-2021-survey,shorten2019survey}.
Furthermore, many researchers develop various data augmentation strategies based on unverified heuristics.
Despite the success of data augmentation, there appears to be a lack of research on why it works.
To explore the effectiveness of these augmentation heuristics,~\citet{kashefi-hwa-2020-quantifying} proposed to quantify an augmentation heuristic for text classification based on the idea that a good heuristic should generate ``hard to distinguish'' samples for different classes.
~\citet{gontijo2020affinity} introduced interpretable and easy-to-compute measures to qualify an augmentation heuristic in the field of Computer Vision~\cite{DBLP:journals/patterns/LiuLTLZ22}.

\paragraph{Pseudo Data Generation.}
Increasing the scale of training data has been instrumental in improving GEC systems.
However, the lack of high-quality publicly available data remains a challenge in low-resource settings~\cite{ma-etal-2022-linguistic, DBLP:journals/corr/abs-2207-08087}.
To mitigate this issue, recent studies focus on generating pseudo data from clean monolingual corpora.
A common approach is to artificially perturb a grammatically correct sentence through random word or character-level insertion, substitution, or deletion~\cite{kiyono2020massive,xu-etal-2019-erroneous,zhou-etal-2020-improving-grammatical}, using spell checkers~\cite{grundkiewicz-etal-2019-neural} or error patterns extracted from realistic datasets~\cite{choe-etal-2019-neural}.
However, these rule-based methods struggle to emulate human-made errors, which can lead to low sample efficiency problem and performance degradation~\cite{yuan-felice-2013-constrained}.
Recent studies utilize models to generate genuine pseudo data, such as backtranslation~\cite{xie-etal-2018-noising} and round-trip translation~\cite{zhou-etal-2020-improving-grammatical}.
On the other hand, \citet{koyama-etal-2021-comparison} shown that the performance of GEC models improves when pseudo datasets generated by various backtranslation models are combined.
~\citet{stahlberg-kumar-2021-synthetic} proposed to generate pseudo data based on a given error type tag using the Seq2Edit model~\cite{stahlberg-kumar-2020-seq2edits}.

\section{Conclusion}\label{sec:conclusion}
This paper introduces two interpretable and computationally efficient measures, Affinity and Diversity, to investigate how data augmentation improves GEC performance.
Our findings demonstrate that an excellent GEC data augmentation strategy characterized by high Affinity and appropriate Diversity can better improve the performance of GEC models.
Inspired by this, we propose MixEdit, which does not require extra monolingual corpora.
Experiments on mainstream datasets in two languages show that MixEdit is effective and complementary to traditional data augmentation methods.

\section*{Limitations}
One shall cautiously consider that our proposed measures, Affinity and Diversity, are a tool for gaining a new perspective on understanding GEC data augmentation.
Though positive correlations are observed between Affinity and performance, it should not be relied upon as a precise predictor for comparing data augmentation methods.
Additionally, it is worth noting that our proposed MixEdit approach is only applicable to realistic datasets, where it can generate label-preserving grammatical errors.
Despite these limitations, we believe that our findings provide a solid foundation for further scientific investigation into GEC data augmentation.

\section*{Ethics Statement}
In this paper, we revisit the effectiveness of traditional data augmentation methods for GEC, including direct noise (DN), pattern noise (PN), backtranslation (BT) and round-trip translation (RT).
The source data for these methods is obtained exclusively from publicly available project resources on legitimate websites, without any involvement of sensitive information.
Additionally, all the baselines and datasets used in our experiments are also publicly accessible, and we have acknowledged the corresponding authors by citing their work.

\section*{Acknowledgements}
This research is supported by National Natural Science Foundation of China (Grant No.62276154), Research  Center for Computer Network (Shenzhen) Ministry of Education, the Natural Science Foundation of Guangdong Province (Grant No.  2023A1515012914), Basic Research Fund of Shenzhen City (Grant No. JCYJ20210324120012033 and JSGG20210802154402007), the Major Key Project of PCL for Experiments and Applications (PCL2021A06), and Overseas Cooperation Research Fund of Tsinghua Shenzhen International
Graduate School (HW2021008).

\bibliography{anthology,custom}

\begin{thebibliography}{60}
\expandafter\ifx\csname natexlab\endcsname\relax\def\natexlab#1{#1}\fi

\bibitem[{Bryant et~al.(2019)Bryant, Felice, Andersen, and
  Briscoe}]{bryant-etal-2019-bea}
Christopher Bryant, Mariano Felice, {\O}istein~E. Andersen, and Ted Briscoe.
  2019.
\newblock \href {https://doi.org/10.18653/v1/W19-4406} {The {BEA}-2019 shared
  task on grammatical error correction}.
\newblock In \emph{Proceedings of the Fourteenth Workshop on Innovative Use of
  NLP for Building Educational Applications}, pages 52--75, Florence, Italy.
  Association for Computational Linguistics.

\bibitem[{Bryant et~al.(2017)Bryant, Felice, and
  Briscoe}]{bryant-etal-2017-automatic}
Christopher Bryant, Mariano Felice, and Ted Briscoe. 2017.
\newblock \href {https://doi.org/10.18653/v1/P17-1074} {Automatic annotation
  and evaluation of error types for grammatical error correction}.
\newblock In \emph{Proceedings of the 55th Annual Meeting of the Association
  for Computational Linguistics (Volume 1: Long Papers)}, pages 793--805,
  Vancouver, Canada. Association for Computational Linguistics.

\bibitem[{Bryant et~al.(2022)Bryant, Yuan, Qorib, Cao, Ng, and
  Briscoe}]{bryant2022grammatical}
Christopher Bryant, Zheng Yuan, Muhammad~Reza Qorib, Hannan Cao, Hwee~Tou Ng,
  and Ted Briscoe. 2022.
\newblock Grammatical error correction: A survey of the state of the art.
\newblock \emph{arXiv preprint arXiv:2211.05166}.

\bibitem[{Chen et~al.(2021)Chen, Shen, Chen, and
  Yang}]{chen-etal-2021-hiddencut}
Jiaao Chen, Dinghan Shen, Weizhu Chen, and Diyi Yang. 2021.
\newblock \href {https://doi.org/10.18653/v1/2021.acl-long.338} {{H}idden{C}ut:
  Simple data augmentation for natural language understanding with better
  generalizability}.
\newblock In \emph{Proceedings of the 59th Annual Meeting of the Association
  for Computational Linguistics and the 11th International Joint Conference on
  Natural Language Processing (Volume 1: Long Papers)}, pages 4380--4390,
  Online. Association for Computational Linguistics.

\bibitem[{Choe et~al.(2019)Choe, Ham, Park, and Yoon}]{choe-etal-2019-neural}
Yo~Joong Choe, Jiyeon Ham, Kyubyong Park, and Yeoil Yoon. 2019.
\newblock \href {https://doi.org/10.18653/v1/W19-4423} {A neural grammatical
  error correction system built on better pre-training and sequential transfer
  learning}.
\newblock In \emph{Proceedings of the Fourteenth Workshop on Innovative Use of
  NLP for Building Educational Applications}, pages 213--227, Florence, Italy.
  Association for Computational Linguistics.

\bibitem[{Dahlmeier and Ng(2012)}]{dahlmeier-ng-2012-better}
Daniel Dahlmeier and Hwee~Tou Ng. 2012.
\newblock \href {https://aclanthology.org/N12-1067} {Better evaluation for
  grammatical error correction}.
\newblock In \emph{Proceedings of the 2012 Conference of the North {A}merican
  Chapter of the Association for Computational Linguistics: Human Language
  Technologies}, pages 568--572, Montr{\'e}al, Canada. Association for
  Computational Linguistics.

\bibitem[{Dahlmeier et~al.(2013)Dahlmeier, Ng, and
  Wu}]{dahlmeier-etal-2013-building}
Daniel Dahlmeier, Hwee~Tou Ng, and Siew~Mei Wu. 2013.
\newblock \href {https://aclanthology.org/W13-1703} {Building a large annotated
  corpus of learner {E}nglish: The {NUS} corpus of learner {E}nglish}.
\newblock In \emph{Proceedings of the Eighth Workshop on Innovative Use of
  {NLP} for Building Educational Applications}, pages 22--31, Atlanta, Georgia.
  Association for Computational Linguistics.

\bibitem[{Dong et~al.(2023)Dong, Li, Gong, Chen, Li, Shen, and
  Yang}]{DBLP:journals/csur/DongLGCLSY23}
Chenhe Dong, Yinghui Li, Haifan Gong, Miaoxin Chen, Junxin Li, Ying Shen, and
  Min Yang. 2023.
\newblock \href {https://doi.org/10.1145/3554727} {A survey of natural language
  generation}.
\newblock \emph{{ACM} Comput. Surv.}, 55(8):173:1--173:38.

\bibitem[{Feng et~al.(2021)Feng, Gangal, Wei, Chandar, Vosoughi, Mitamura, and
  Hovy}]{feng-etal-2021-survey}
Steven~Y. Feng, Varun Gangal, Jason Wei, Sarath Chandar, Soroush Vosoughi,
  Teruko Mitamura, and Eduard Hovy. 2021.
\newblock \href {https://doi.org/10.18653/v1/2021.findings-acl.84} {A survey of
  data augmentation approaches for {NLP}}.
\newblock In \emph{Findings of the Association for Computational Linguistics:
  ACL-IJCNLP 2021}, pages 968--988, Online. Association for Computational
  Linguistics.

\bibitem[{Gontijo-Lopes et~al.(2020)Gontijo-Lopes, Smullin, Cubuk, and
  Dyer}]{gontijo2020affinity}
Raphael Gontijo-Lopes, Sylvia~J Smullin, Ekin~D Cubuk, and Ethan Dyer. 2020.
\newblock Affinity and diversity: Quantifying mechanisms of data augmentation.
\newblock \emph{arXiv preprint arXiv:2002.08973}.

\bibitem[{Grundkiewicz and
  Junczys-Dowmunt(2019)}]{grundkiewicz-junczys-dowmunt-2019-minimally}
Roman Grundkiewicz and Marcin Junczys-Dowmunt. 2019.
\newblock \href {https://doi.org/10.18653/v1/D19-5546} {Minimally-augmented
  grammatical error correction}.
\newblock In \emph{Proceedings of the 5th Workshop on Noisy User-generated Text
  (W-NUT 2019)}, pages 357--363, Hong Kong, China. Association for
  Computational Linguistics.

\bibitem[{Grundkiewicz et~al.(2019)Grundkiewicz, Junczys-Dowmunt, and
  Heafield}]{grundkiewicz-etal-2019-neural}
Roman Grundkiewicz, Marcin Junczys-Dowmunt, and Kenneth Heafield. 2019.
\newblock \href {https://doi.org/10.18653/v1/W19-4427} {Neural grammatical
  error correction systems with unsupervised pre-training on synthetic data}.
\newblock In \emph{Proceedings of the Fourteenth Workshop on Innovative Use of
  NLP for Building Educational Applications}, pages 252--263, Florence, Italy.
  Association for Computational Linguistics.

\bibitem[{Junczys-Dowmunt et~al.(2018)Junczys-Dowmunt, Grundkiewicz, Guha, and
  Heafield}]{junczys-dowmunt-etal-2018-approaching}
Marcin Junczys-Dowmunt, Roman Grundkiewicz, Shubha Guha, and Kenneth Heafield.
  2018.
\newblock \href {https://doi.org/10.18653/v1/N18-1055} {Approaching neural
  grammatical error correction as a low-resource machine translation task}.
\newblock In \emph{Proceedings of the 2018 Conference of the North {A}merican
  Chapter of the Association for Computational Linguistics: Human Language
  Technologies, Volume 1 (Long Papers)}, pages 595--606, New Orleans,
  Louisiana. Association for Computational Linguistics.

\bibitem[{Kaneko et~al.(2020)Kaneko, Mita, Kiyono, Suzuki, and
  Inui}]{kaneko-etal-2020-encoder}
Masahiro Kaneko, Masato Mita, Shun Kiyono, Jun Suzuki, and Kentaro Inui. 2020.
\newblock \href {https://doi.org/10.18653/v1/2020.acl-main.391}
  {Encoder-decoder models can benefit from pre-trained masked language models
  in grammatical error correction}.
\newblock In \emph{Proceedings of the 58th Annual Meeting of the Association
  for Computational Linguistics}, pages 4248--4254, Online. Association for
  Computational Linguistics.

\bibitem[{Kashefi and Hwa(2020)}]{kashefi-hwa-2020-quantifying}
Omid Kashefi and Rebecca Hwa. 2020.
\newblock \href {https://doi.org/10.18653/v1/2020.wnut-1.26} {Quantifying the
  evaluation of heuristic methods for textual data augmentation}.
\newblock In \emph{Proceedings of the Sixth Workshop on Noisy User-generated
  Text (W-NUT 2020)}, pages 200--208, Online. Association for Computational
  Linguistics.

\bibitem[{Katsumata and Komachi(2020)}]{katsumata-komachi-2020-stronger}
Satoru Katsumata and Mamoru Komachi. 2020.
\newblock \href {https://aclanthology.org/2020.aacl-main.83} {Stronger
  baselines for grammatical error correction using a pretrained encoder-decoder
  model}.
\newblock In \emph{Proceedings of the 1st Conference of the Asia-Pacific
  Chapter of the Association for Computational Linguistics and the 10th
  International Joint Conference on Natural Language Processing}, pages
  827--832, Suzhou, China. Association for Computational Linguistics.

\bibitem[{Kingma and Ba(2014)}]{kingma2014adam}
Diederik~P Kingma and Jimmy Ba. 2014.
\newblock Adam: A method for stochastic optimization.
\newblock \emph{arXiv preprint arXiv:1412.6980}.

\bibitem[{Kiyono et~al.(2019)Kiyono, Suzuki, Mita, Mizumoto, and
  Inui}]{kiyono-etal-2019-empirical}
Shun Kiyono, Jun Suzuki, Masato Mita, Tomoya Mizumoto, and Kentaro Inui. 2019.
\newblock \href {https://doi.org/10.18653/v1/D19-1119} {An empirical study of
  incorporating pseudo data into grammatical error correction}.
\newblock In \emph{Proceedings of the 2019 Conference on Empirical Methods in
  Natural Language Processing and the 9th International Joint Conference on
  Natural Language Processing (EMNLP-IJCNLP)}, pages 1236--1242, Hong Kong,
  China. Association for Computational Linguistics.

\bibitem[{Kiyono et~al.(2020)Kiyono, Suzuki, Mizumoto, and
  Inui}]{kiyono2020massive}
Shun Kiyono, Jun Suzuki, Tomoya Mizumoto, and Kentaro Inui. 2020.
\newblock Massive exploration of pseudo data for grammatical error correction.
\newblock \emph{IEEE/ACM transactions on audio, speech, and language
  processing}, 28:2134--2145.

\bibitem[{Koyama et~al.(2021{\natexlab{a}})Koyama, Hotate, Kaneko, and
  Komachi}]{koyama-etal-2021-comparison}
Aomi Koyama, Kengo Hotate, Masahiro Kaneko, and Mamoru Komachi.
  2021{\natexlab{a}}.
\newblock \href {https://doi.org/10.18653/v1/2021.naacl-srw.16} {Comparison of
  grammatical error correction using back-translation models}.
\newblock In \emph{Proceedings of the 2021 Conference of the North American
  Chapter of the Association for Computational Linguistics: Student Research
  Workshop}, pages 126--135, Online. Association for Computational Linguistics.

\bibitem[{Koyama et~al.(2021{\natexlab{b}})Koyama, Takamura, and
  Okazaki}]{koyama2021various}
Shota Koyama, Hiroya Takamura, and Naoaki Okazaki. 2021{\natexlab{b}}.
\newblock Various errors improve neural grammatical error correction.
\newblock In \emph{Proceedings of the 35th Pacific Asia Conference on Language,
  Information and Computation}, pages 251--261.

\bibitem[{Lewis et~al.(2020)Lewis, Liu, Goyal, Ghazvininejad, Mohamed, Levy,
  Stoyanov, and Zettlemoyer}]{lewis-etal-2020-bart}
Mike Lewis, Yinhan Liu, Naman Goyal, Marjan Ghazvininejad, Abdelrahman Mohamed,
  Omer Levy, Veselin Stoyanov, and Luke Zettlemoyer. 2020.
\newblock \href {https://doi.org/10.18653/v1/2020.acl-main.703} {{BART}:
  Denoising sequence-to-sequence pre-training for natural language generation,
  translation, and comprehension}.
\newblock In \emph{Proceedings of the 58th Annual Meeting of the Association
  for Computational Linguistics}, pages 7871--7880, Online. Association for
  Computational Linguistics.

\bibitem[{Li et~al.(2023)Li, Huang, Ma, Jiang, Li, Zhou, Zheng, and
  Zhou}]{DBLP:journals/corr/abs-2307-09007}
Yinghui Li, Haojing Huang, Shirong Ma, Yong Jiang, Yangning Li, Feng Zhou,
  Hai{-}Tao Zheng, and Qingyu Zhou. 2023.
\newblock \href {https://doi.org/10.48550/arXiv.2307.09007} {On the
  (in)effectiveness of large language models for chinese text correction}.
\newblock \emph{CoRR}, abs/2307.09007.

\bibitem[{Li et~al.(2022{\natexlab{a}})Li, Huang, Zhang, Zhou, Li, Liu, Cao,
  Zheng, and Shen}]{DBLP:journals/corr/abs-2207-08087}
Yinghui Li, Shulin Huang, Xinwei Zhang, Qingyu Zhou, Yangning Li, Ruiyang Liu,
  Yunbo Cao, Hai{-}Tao Zheng, and Ying Shen. 2022{\natexlab{a}}.
\newblock \href {https://doi.org/10.48550/arXiv.2207.08087} {Automatic context
  pattern generation for entity set expansion}.
\newblock \emph{CoRR}, abs/2207.08087.

\bibitem[{Li et~al.(2022{\natexlab{b}})Li, Ma, Zhou, Li, Li, Huang, Liu, Li,
  Cao, and Zheng}]{DBLP:conf/emnlp/LiMZLLHLLC022}
Yinghui Li, Shirong Ma, Qingyu Zhou, Zhongli Li, Yangning Li, Shulin Huang,
  Ruiyang Liu, Chao Li, Yunbo Cao, and Haitao Zheng. 2022{\natexlab{b}}.
\newblock \href {https://doi.org/10.18653/v1/2022.findings-emnlp.18} {Learning
  from the dictionary: Heterogeneous knowledge guided fine-tuning for chinese
  spell checking}.
\newblock In \emph{Findings of the Association for Computational Linguistics:
  {EMNLP} 2022, Abu Dhabi, United Arab Emirates, December 7-11, 2022}, pages
  238--249. Association for Computational Linguistics.

\bibitem[{Li et~al.(2022{\natexlab{c}})Li, Zhou, Li, Li, Liu, Sun, Wang, Li,
  Cao, and Zheng}]{DBLP:conf/acl/LiZLLLSWLCZ22}
Yinghui Li, Qingyu Zhou, Yangning Li, Zhongli Li, Ruiyang Liu, Rongyi Sun,
  Zizhen Wang, Chao Li, Yunbo Cao, and Hai{-}Tao Zheng. 2022{\natexlab{c}}.
\newblock \href {https://doi.org/10.18653/v1/2022.findings-acl.252} {The past
  mistake is the future wisdom: Error-driven contrastive probability
  optimization for chinese spell checking}.
\newblock In \emph{Findings of the Association for Computational Linguistics:
  {ACL} 2022, Dublin, Ireland, May 22-27, 2022}, pages 3202--3213. Association
  for Computational Linguistics.

\bibitem[{Lichtarge et~al.(2020)Lichtarge, Alberti, and
  Kumar}]{lichtarge-etal-2020-data}
Jared Lichtarge, Chris Alberti, and Shankar Kumar. 2020.
\newblock \href {https://doi.org/10.1162/tacl_a_00336} {Data weighted training
  strategies for grammatical error correction}.
\newblock \emph{Transactions of the Association for Computational Linguistics},
  8:634--646.

\bibitem[{Lichtarge et~al.(2019)Lichtarge, Alberti, Kumar, Shazeer, Parmar, and
  Tong}]{lichtarge-etal-2019-corpora}
Jared Lichtarge, Chris Alberti, Shankar Kumar, Noam Shazeer, Niki Parmar, and
  Simon Tong. 2019.
\newblock \href {https://doi.org/10.18653/v1/N19-1333} {Corpora generation for
  grammatical error correction}.
\newblock In \emph{Proceedings of the 2019 Conference of the North {A}merican
  Chapter of the Association for Computational Linguistics: Human Language
  Technologies, Volume 1 (Long and Short Papers)}, pages 3291--3301,
  Minneapolis, Minnesota. Association for Computational Linguistics.

\bibitem[{Liu et~al.(2022)Liu, Li, Tao, Liang, and
  Zheng}]{DBLP:journals/patterns/LiuLTLZ22}
Ruiyang Liu, Yinghui Li, Linmi Tao, Dun Liang, and Hai{-}Tao Zheng. 2022.
\newblock \href {https://doi.org/10.1016/j.patter.2022.100520} {Are we ready
  for a new paradigm shift? {A} survey on visual deep {MLP}}.
\newblock \emph{Patterns}, 3(7):100520.

\bibitem[{Ma et~al.(2023)Ma, Li, Huang, Huang, Li, Zheng, and
  Shen}]{DBLP:journals/corr/abs-2306-17447}
Shirong Ma, Yinghui Li, Haojing Huang, Shulin Huang, Yangning Li, Hai{-}Tao
  Zheng, and Ying Shen. 2023.
\newblock \href {https://doi.org/10.48550/arXiv.2306.17447} {Progressive
  multi-task learning framework for chinese text error correction}.
\newblock \emph{CoRR}, abs/2306.17447.

\bibitem[{Ma et~al.(2022)Ma, Li, Sun, Zhou, Huang, Zhang, Yangning, Liu, Li,
  Cao, Zheng, and Shen}]{ma-etal-2022-linguistic}
Shirong Ma, Yinghui Li, Rongyi Sun, Qingyu Zhou, Shulin Huang, Ding Zhang,
  Li~Yangning, Ruiyang Liu, Zhongli Li, Yunbo Cao, Haitao Zheng, and Ying Shen.
  2022.
\newblock \href {https://aclanthology.org/2022.findings-emnlp.40} {Linguistic
  rules-based corpus generation for native {C}hinese grammatical error
  correction}.
\newblock In \emph{Findings of the Association for Computational Linguistics:
  EMNLP 2022}, pages 576--589, Abu Dhabi, United Arab Emirates. Association for
  Computational Linguistics.

\bibitem[{Ng et~al.(2014)Ng, Wu, Briscoe, Hadiwinoto, Susanto, and
  Bryant}]{ng-etal-2014-conll}
Hwee~Tou Ng, Siew~Mei Wu, Ted Briscoe, Christian Hadiwinoto, Raymond~Hendy
  Susanto, and Christopher Bryant. 2014.
\newblock \href {https://doi.org/10.3115/v1/W14-1701} {The {C}o{NLL}-2014
  shared task on grammatical error correction}.
\newblock In \emph{Proceedings of the Eighteenth Conference on Computational
  Natural Language Learning: Shared Task}, pages 1--14, Baltimore, Maryland.
  Association for Computational Linguistics.

\bibitem[{Omelianchuk et~al.(2020)Omelianchuk, Atrasevych, Chernodub, and
  Skurzhanskyi}]{omelianchuk-etal-2020-gector}
Kostiantyn Omelianchuk, Vitaliy Atrasevych, Artem Chernodub, and Oleksandr
  Skurzhanskyi. 2020.
\newblock \href {https://doi.org/10.18653/v1/2020.bea-1.16} {{GECT}o{R} {--}
  grammatical error correction: Tag, not rewrite}.
\newblock In \emph{Proceedings of the Fifteenth Workshop on Innovative Use of
  NLP for Building Educational Applications}, pages 163--170, Seattle, WA, USA
  → Online. Association for Computational Linguistics.

\bibitem[{Ott et~al.(2019)Ott, Edunov, Baevski, Fan, Gross, Ng, Grangier, and
  Auli}]{ott-etal-2019-fairseq}
Myle Ott, Sergey Edunov, Alexei Baevski, Angela Fan, Sam Gross, Nathan Ng,
  David Grangier, and Michael Auli. 2019.
\newblock \href {https://doi.org/10.18653/v1/N19-4009} {fairseq: A fast,
  extensible toolkit for sequence modeling}.
\newblock In \emph{Proceedings of the 2019 Conference of the North {A}merican
  Chapter of the Association for Computational Linguistics (Demonstrations)},
  pages 48--53, Minneapolis, Minnesota. Association for Computational
  Linguistics.

\bibitem[{Rothe et~al.(2021)Rothe, Mallinson, Malmi, Krause, and
  Severyn}]{rothe-etal-2021-simple}
Sascha Rothe, Jonathan Mallinson, Eric Malmi, Sebastian Krause, and Aliaksei
  Severyn. 2021.
\newblock \href {https://doi.org/10.18653/v1/2021.acl-short.89} {A simple
  recipe for multilingual grammatical error correction}.
\newblock In \emph{Proceedings of the 59th Annual Meeting of the Association
  for Computational Linguistics and the 11th International Joint Conference on
  Natural Language Processing (Volume 2: Short Papers)}, pages 702--707,
  Online. Association for Computational Linguistics.

\bibitem[{Sennrich et~al.(2016{\natexlab{a}})Sennrich, Haddow, and
  Birch}]{sennrich-etal-2016-improving}
Rico Sennrich, Barry Haddow, and Alexandra Birch. 2016{\natexlab{a}}.
\newblock \href {https://doi.org/10.18653/v1/P16-1009} {Improving neural
  machine translation models with monolingual data}.
\newblock In \emph{Proceedings of the 54th Annual Meeting of the Association
  for Computational Linguistics (Volume 1: Long Papers)}, pages 86--96, Berlin,
  Germany. Association for Computational Linguistics.

\bibitem[{Sennrich et~al.(2016{\natexlab{b}})Sennrich, Haddow, and
  Birch}]{sennrich-etal-2016-neural}
Rico Sennrich, Barry Haddow, and Alexandra Birch. 2016{\natexlab{b}}.
\newblock \href {https://doi.org/10.18653/v1/P16-1162} {Neural machine
  translation of rare words with subword units}.
\newblock In \emph{Proceedings of the 54th Annual Meeting of the Association
  for Computational Linguistics (Volume 1: Long Papers)}, pages 1715--1725,
  Berlin, Germany. Association for Computational Linguistics.

\bibitem[{Shao et~al.(2021)Shao, Geng, Liu, Dai, Yan, Yang, Zhe, Bao, and
  Qiu}]{shao2021cpt}
Yunfan Shao, Zhichao Geng, Yitao Liu, Junqi Dai, Hang Yan, Fei Yang, Li~Zhe,
  Hujun Bao, and Xipeng Qiu. 2021.
\newblock Cpt: A pre-trained unbalanced transformer for both chinese language
  understanding and generation.
\newblock \emph{arXiv preprint arXiv:2109.05729}.

\bibitem[{Shen et~al.(2020)Shen, Zheng, Shen, Qu, and Chen}]{shen2020simple}
Dinghan Shen, Mingzhi Zheng, Yelong Shen, Yanru Qu, and Weizhu Chen. 2020.
\newblock A simple but tough-to-beat data augmentation approach for natural
  language understanding and generation.
\newblock \emph{arXiv preprint arXiv:2009.13818}.

\bibitem[{Shorten and Khoshgoftaar(2019)}]{shorten2019survey}
Connor Shorten and Taghi~M Khoshgoftaar. 2019.
\newblock A survey on image data augmentation for deep learning.
\newblock \emph{Journal of big data}, 6(1):1--48.

\bibitem[{Stahlberg and Kumar(2020)}]{stahlberg-kumar-2020-seq2edits}
Felix Stahlberg and Shankar Kumar. 2020.
\newblock \href {https://doi.org/10.18653/v1/2020.emnlp-main.418}
  {{S}eq2{E}dits: Sequence transduction using span-level edit operations}.
\newblock In \emph{Proceedings of the 2020 Conference on Empirical Methods in
  Natural Language Processing (EMNLP)}, pages 5147--5159, Online. Association
  for Computational Linguistics.

\bibitem[{Stahlberg and Kumar(2021)}]{stahlberg-kumar-2021-synthetic}
Felix Stahlberg and Shankar Kumar. 2021.
\newblock \href {https://aclanthology.org/2021.bea-1.4} {Synthetic data
  generation for grammatical error correction with tagged corruption models}.
\newblock In \emph{Proceedings of the 16th Workshop on Innovative Use of NLP
  for Building Educational Applications}, pages 37--47, Online. Association for
  Computational Linguistics.

\bibitem[{Sun et~al.(2021)Sun, Ge, Wei, and Wang}]{sun-etal-2021-instantaneous}
Xin Sun, Tao Ge, Furu Wei, and Houfeng Wang. 2021.
\newblock \href {https://doi.org/10.18653/v1/2021.acl-long.462} {Instantaneous
  grammatical error correction with shallow aggressive decoding}.
\newblock In \emph{Proceedings of the 59th Annual Meeting of the Association
  for Computational Linguistics and the 11th International Joint Conference on
  Natural Language Processing (Volume 1: Long Papers)}, pages 5937--5947,
  Online. Association for Computational Linguistics.

\bibitem[{Tiedemann and Thottingal(2020)}]{TiedemannThottingal:EAMT2020}
J{\"o}rg Tiedemann and Santhosh Thottingal. 2020.
\newblock {OPUS-MT} — {B}uilding open translation services for the {W}orld.
\newblock In \emph{Proceedings of the 22nd Annual Conferenec of the European
  Association for Machine Translation (EAMT)}, Lisbon, Portugal.

\bibitem[{Tu et~al.(2020)Tu, Lalwani, Gella, and He}]{tu-etal-2020-empirical}
Lifu Tu, Garima Lalwani, Spandana Gella, and He~He. 2020.
\newblock \href {https://doi.org/10.1162/tacl_a_00335} {An empirical study on
  robustness to spurious correlations using pre-trained language models}.
\newblock \emph{Transactions of the Association for Computational Linguistics},
  8:621--633.

\bibitem[{White and Rozovskaya(2020)}]{white-rozovskaya-2020-comparative}
Max White and Alla Rozovskaya. 2020.
\newblock \href {https://doi.org/10.18653/v1/2020.bea-1.21} {A comparative
  study of synthetic data generation methods for grammatical error correction}.
\newblock In \emph{Proceedings of the Fifteenth Workshop on Innovative Use of
  NLP for Building Educational Applications}, pages 198--208, Seattle, WA, USA
  → Online. Association for Computational Linguistics.

\bibitem[{Xie et~al.(2018)Xie, Genthial, Xie, Ng, and
  Jurafsky}]{xie-etal-2018-noising}
Ziang Xie, Guillaume Genthial, Stanley Xie, Andrew Ng, and Dan Jurafsky. 2018.
\newblock \href {https://doi.org/10.18653/v1/N18-1057} {Noising and denoising
  natural language: Diverse backtranslation for grammar correction}.
\newblock In \emph{Proceedings of the 2018 Conference of the North {A}merican
  Chapter of the Association for Computational Linguistics: Human Language
  Technologies, Volume 1 (Long Papers)}, pages 619--628, New Orleans,
  Louisiana. Association for Computational Linguistics.

\bibitem[{Xu et~al.(2019)Xu, Zhang, Chen, and Qin}]{xu-etal-2019-erroneous}
Shuyao Xu, Jiehao Zhang, Jin Chen, and Long Qin. 2019.
\newblock \href {https://doi.org/10.18653/v1/W19-4415} {Erroneous data
  generation for grammatical error correction}.
\newblock In \emph{Proceedings of the Fourteenth Workshop on Innovative Use of
  NLP for Building Educational Applications}, pages 149--158, Florence, Italy.
  Association for Computational Linguistics.

\bibitem[{Yannakoudakis et~al.(2011)Yannakoudakis, Briscoe, and
  Medlock}]{yannakoudakis-etal-2011-new}
Helen Yannakoudakis, Ted Briscoe, and Ben Medlock. 2011.
\newblock \href {https://aclanthology.org/P11-1019} {A new dataset and method
  for automatically grading {ESOL} texts}.
\newblock In \emph{Proceedings of the 49th Annual Meeting of the Association
  for Computational Linguistics: Human Language Technologies}, pages 180--189,
  Portland, Oregon, USA. Association for Computational Linguistics.

\bibitem[{Yasunaga et~al.(2021)Yasunaga, Leskovec, and
  Liang}]{yasunaga-etal-2021-lm}
Michihiro Yasunaga, Jure Leskovec, and Percy Liang. 2021.
\newblock \href {https://doi.org/10.18653/v1/2021.emnlp-main.611} {{LM}-critic:
  Language models for unsupervised grammatical error correction}.
\newblock In \emph{Proceedings of the 2021 Conference on Empirical Methods in
  Natural Language Processing}, pages 7752--7763, Online and Punta Cana,
  Dominican Republic. Association for Computational Linguistics.

\bibitem[{Ye et~al.(2022)Ye, Li, Ma, Xie, Wu, and
  Zheng}]{DBLP:journals/corr/abs-2210-12692}
Jingheng Ye, Yinghui Li, Shirong Ma, Rui Xie, Wei Wu, and Hai{-}Tao Zheng.
  2022.
\newblock \href {https://doi.org/10.48550/arXiv.2210.12692} {Focus is what you
  need for chinese grammatical error correction}.
\newblock \emph{CoRR}, abs/2210.12692.

\bibitem[{Ye et~al.(2023{\natexlab{a}})Ye, Li, and Zheng}]{ye-etal-2023-system}
Jingheng Ye, Yinghui Li, and Haitao Zheng. 2023{\natexlab{a}}.
\newblock \href {https://aclanthology.org/2023.ccl-3.29} {System report for
  {CCL}23-eval task 7: {THU} {KEL}ab (sz) - exploring data augmentation and
  denoising for {C}hinese grammatical error correction}.
\newblock In \emph{Proceedings of the 22nd Chinese National Conference on
  Computational Linguistics (Volume 3: Evaluations)}, pages 262--270, Harbin,
  China. Chinese Information Processing Society of China.

\bibitem[{Ye et~al.(2023{\natexlab{b}})Ye, Li, Zhou, Li, Ma, Zheng, and
  Shen}]{DBLP:journals/corr/abs-2305-10819}
Jingheng Ye, Yinghui Li, Qingyu Zhou, Yangning Li, Shirong Ma, Hai{-}Tao Zheng,
  and Ying Shen. 2023{\natexlab{b}}.
\newblock \href {https://doi.org/10.48550/arXiv.2305.10819} {{CLEME:} debiasing
  multi-reference evaluation for grammatical error correction}.
\newblock \emph{CoRR}, abs/2305.10819.

\bibitem[{Yuan and Felice(2013)}]{yuan-felice-2013-constrained}
Zheng Yuan and Mariano Felice. 2013.
\newblock \href {https://aclanthology.org/W13-3607} {Constrained grammatical
  error correction using statistical machine translation}.
\newblock In \emph{Proceedings of the Seventeenth Conference on Computational
  Natural Language Learning: Shared Task}, pages 52--61, Sofia, Bulgaria.
  Association for Computational Linguistics.

\bibitem[{Zhang(2009)}]{zhang2009features}
Baolin Zhang. 2009.
\newblock Features and functions of the hsk dynamic composition corpus.
\newblock \emph{International Chinese Language Education}, 4:71--79.

\bibitem[{Zhang et~al.(2023)Zhang, Li, Zhou, Ma, Li, Cao, and Zheng}]{10095675}
Ding Zhang, Yinghui Li, Qingyu Zhou, Shirong Ma, Yangning Li, Yunbo Cao, and
  Hai-Tao Zheng. 2023.
\newblock \href {https://doi.org/10.1109/ICASSP49357.2023.10095675} {Contextual
  similarity is more valuable than character similarity: An empirical study for
  chinese spell checking}.
\newblock In \emph{ICASSP 2023 - 2023 IEEE International Conference on
  Acoustics, Speech and Signal Processing (ICASSP)}, pages 1--5.

\bibitem[{Zhang et~al.(2022{\natexlab{a}})Zhang, Li, Bao, Li, Zhang, Li, Huang,
  and Zhang}]{zhang-etal-2022-mucgec}
Yue Zhang, Zhenghua Li, Zuyi Bao, Jiacheng Li, Bo~Zhang, Chen Li, Fei Huang,
  and Min Zhang. 2022{\natexlab{a}}.
\newblock \href {https://doi.org/10.18653/v1/2022.naacl-main.227} {{M}u{CGEC}:
  a multi-reference multi-source evaluation dataset for {C}hinese grammatical
  error correction}.
\newblock In \emph{Proceedings of the 2022 Conference of the North American
  Chapter of the Association for Computational Linguistics: Human Language
  Technologies}, pages 3118--3130, Seattle, United States. Association for
  Computational Linguistics.

\bibitem[{Zhang et~al.(2022{\natexlab{b}})Zhang, Zhang, Li, Bao, Li, and
  Zhang}]{zhang-etal-2022-syngec}
Yue Zhang, Bo~Zhang, Zhenghua Li, Zuyi Bao, Chen Li, and Min Zhang.
  2022{\natexlab{b}}.
\newblock \href {https://aclanthology.org/2022.emnlp-main.162} {{S}yn{GEC}:
  Syntax-enhanced grammatical error correction with a tailored {GEC}-oriented
  parser}.
\newblock In \emph{Proceedings of the 2022 Conference on Empirical Methods in
  Natural Language Processing}, pages 2518--2531, Abu Dhabi, United Arab
  Emirates. Association for Computational Linguistics.

\bibitem[{Zhao et~al.(2018)Zhao, Jiang, Sun, and
  Wan}]{10.1007/978-3-319-99501-4_41}
Yuanyuan Zhao, Nan Jiang, Weiwei Sun, and Xiaojun Wan. 2018.
\newblock Overview of the nlpcc 2018 shared task: Grammatical error correction.
\newblock In \emph{Natural Language Processing and Chinese Computing}, pages
  439--445, Cham. Springer International Publishing.

\bibitem[{Zhou et~al.(2020)Zhou, Ge, Mu, Xu, Wei, and
  Zhou}]{zhou-etal-2020-improving-grammatical}
Wangchunshu Zhou, Tao Ge, Chang Mu, Ke~Xu, Furu Wei, and Ming Zhou. 2020.
\newblock \href {https://doi.org/10.18653/v1/2020.findings-emnlp.30} {Improving
  grammatical error correction with machine translation pairs}.
\newblock In \emph{Findings of the Association for Computational Linguistics:
  EMNLP 2020}, pages 318--328, Online. Association for Computational
  Linguistics.

\end{thebibliography}
\bibliographystyle{acl_natbib}

\appendix
\section{Implementation Details}\label{appendix:implementation_details}

\subsection{Hyper-parameters}\label{appendix:hyper_parameters}

\begin{table}[th!]
\centering
\scalebox{0.74}{
\begin{tabular}{lc}
\hline
\textbf{Configuration}  & \textbf{Value}    \\ 
\hline

\multicolumn{2}{c}{\textbf{Pre-training}}   \\ \hline
Backbone  &  BART-large \citep{lewis-etal-2020-bart}    \\
Devices   &  4 Tesla A100 GPU (80GB)      \\
Epochs    &  60                           \\
Batch size per GPU & 4096 tokens                        \\

Optimizer & \begin{tabular}[c]{@{}c@{}}
Adam \citep{kingma2014adam}  \\ 
($\beta_1=0.9,\beta_2=0.999,\epsilon=1 \times 10^{-8}$) 
\end{tabular}    \\

Learning rate      &  $3 \times 10^{-5}$                          \\
Warmup updates     & 2000                           \\
Max source length  & 1024                           \\
Dropout            & 0.3 (English); 0.2 (Chinese)                          \\ 
Dropout-src        & 0.2                            \\ 
\hline

\multicolumn{2}{c}{\textbf{Fine-tuning}}                   \\ \hline
Weights of Loss    & $\alpha$=1.0, $\beta$=1.0  \\
Learning rate      & 
\begin{tabular}[c]{@{}c@{}}
$3 \times 10^{-5}$, $5 \times 10^{-6}$, $3 \times 10^{-6}$ (English)        \\
$3 \times 10^{-5}$ (Chinese)                                                \\
\end{tabular}                                       \\
Warmup updates     & 2000                           \\

\hline

\multicolumn{2}{c}{\textbf{Inference}}              \\ 
\hline

Beam size          & 12                             \\

\hline

\end{tabular}}
\caption{Hyper-parameter values used in our experiments.}
\label{tab:hp}
\end{table}

We list the main hyper-parameters in Table~\ref{tab:hp}.
For the pre-training stage, we follow the same hyper-parameters as described in~\cite{zhang-etal-2022-syngec}.
We introduce MixEdit throughout the three-stage fine-tuning.
To determine the optimal balancing weights for the training objective, we experiment with various values for $\alpha$
within \{0.0, 0.4, 0.8, 1.0\} and $\beta$ within \{0.5, 1.0, 2.0\}.
Our experiments reveal that the configuration with $\alpha=1.0$ and $\beta=1.0$ achieves the highest F$_{0.5}$ score on BEA-19 and CoNLL-14, which is used as the default settings in all our experiments.

\subsection{Details of GEC Data Augmentation}\label{appendix:details_augmentation}
We follow the default experimental settings of GEC data augmentation methods as proposed in their papers or source code.
In our experiments, we set $\mu_{\operatorname{mask}}=0.3$ for DN, which has been proven to be the best by~\citet{kiyono2020massive}.
For PN, we generate pseudo data by following the default instructions provided in the open-source project~\footnote{
\url{https://github.com/kakaobrain/helo-word}
}.
We train the BT model, which is initialized with the weights of BART-large, on the Clang8 dataset, setting $\beta_{\operatorname{random}}=6$ to yields the best performance as evidenced in Section~\ref{subsec:analysis}.
As for RT, we generate pseudo data via Chinese as the bridge language, leveraging two off-the-shelf en-zh~\footnote{
\url{https://huggingface.co/Helsinki-NLP/opus-mt-en-zh}
} and zh-en~\footnote{
\url{https://huggingface.co/Helsinki-NLP/opus-mt-zh-en}
} translation models~\cite{TiedemannThottingal:EAMT2020}.

\section{Extra Experiments}

\subsection{Affinity and Diversity of Extra Datasets}\label{app:extra_datasets}

\begin{table}[thbp!]
\centering
\scalebox{0.7}{
\begin{tabular}{lllccc}
\toprule

\textbf{Method}  &  \textbf{Affinity}$^\uparrow$  &  \textbf{Diversity}  &
\textbf{P}  &  \textbf{R} & $\mbox{\textbf{F}}_{0.5}$           \\
\hline

\textbf{Baseline}   &  -     &  8.81   &  \bf{63.04}  &  44.69  &  \textbf{58.25}     \\
\textbf{DN}  &  0.33  &  11.31  &  28.52  &  24.48  &  27.60     \\
\textbf{RT}  &  0.53  &  11.14  &  18.43  &  37.16  &  20.50     \\

\textbf{PN}   \\
\hspace{0.3cm} \textbf{Round=1}  &  \bf{3.74}  &  7.63  &  \bf{62.66}  &  28.32  &  50.43     \\
\hspace{0.3cm} \textbf{Round=2}  &  3.55  &  7.79  &  60.24  &  34.12  &  \bf{52.24}     \\
\hspace{0.3cm} \textbf{Round=4}  &  3.17  &  8.05  &  56.70  &  38.58  &  51.83     \\
\hspace{0.3cm} \textbf{Round=8}  &  2.65  &  8.47  &  49.70  &  43.36  &  48.29     \\
\hspace{0.3cm} \textbf{Round=16} &  1.92  &  \bf{9.10}  &  40.59  &  \bf{48.21}  &  41.92     \\

\textbf{BT}   \\
\hspace{0.3cm} \textbf{$\beta_{\operatorname{random}}$=0}
&  1.04  &  7.04  &  57.50  &  13.96  &  35.41      \\
\hspace{0.3cm} \textbf{$\beta_{\operatorname{random}}$=3}
&  3.59  &  7.60  &  \bf{62.63}  &  38.94  &  \bf{55.83}      \\
\hspace{0.3cm} \textbf{$\beta_{\operatorname{random}}$=6}
&  \bf{3.73}  &  8.15  &  57.73  &  46.71  &  55.13      \\
\hspace{0.3cm} \textbf{$\beta_{\operatorname{random}}$=9}
&  3.08  &  8.55  &  53.07  &  51.19  &  52.68      \\
\hspace{0.3cm} \textbf{$\beta_{\operatorname{random}}$=12}
&  2.55  &  \bf{8.87}  &  49.38  &  \bf{52.63}  &  50.00      \\

\textbf{MixEdit} (Ours)  &  \bf{19.29}  &  8.85   &  \bf{64.50}  &  38.32  &  \textbf{56.75}      \\

\bottomrule
\end{tabular}}

\caption{
Affinity and Diversity of data augmentation methods.
All the models are trained using realistic or pseudo English CLang-8 dataset.
}
\label{tab:affinity_diversity_clang8}

\end{table}
\begin{table}[thbp!]
\centering
\scalebox{0.7}{
\begin{tabular}{lllccc}
\toprule

\textbf{Method}  &  \textbf{Affinity}$^\uparrow$  &  \textbf{Diversity}  &
\textbf{P}  &  \textbf{R} & $\mbox{\textbf{F}}_{0.5}$           \\
\hline

\textbf{Baseline}   &  -     &  8.78   &  \bf{49.10}  &  25.30  &  \bf{41.33}     \\
\textbf{DN}  &  0.39  &  10.10  &  28.84  &  26.56  &  28.35     \\
\textbf{RT}  &  0.46  &  10.57  &  19.39  &  29.84  &  20.84     \\

\textbf{PN}   \\
\hspace{0.3cm} \textbf{Round=1}  &  \bf{3.07}  &  6.40  &  \bf{45.84}  &  08.86  &  24.98     \\
\hspace{0.3cm} \textbf{Round=2}  &  2.73  &  6.59  &  43.54  &  13.28  &  29.91     \\
\hspace{0.3cm} \textbf{Round=4}  &  2.25  &  6.91  &  44.96  &  15.12  &  32.24     \\
\hspace{0.3cm} \textbf{Round=8}  &  1.43  &  7.43  &  42.83  &  16.89  &  32.76     \\
\hspace{0.3cm} \textbf{Round=16} &  1.17  &  \bf{8.20}  &  39.82  &  \bf{21.24}  &  \bf{33.89}     \\

\textbf{BT}   \\
\hspace{0.3cm} \textbf{$\beta_{\operatorname{random}}$=0}
&  1.30  &  5.44  &  27.07  &  03.34  &  11.18      \\
\hspace{0.3cm} \textbf{$\beta_{\operatorname{random}}$=3}
&  \bf{3.00}  &  7.24  &  \bf{45.48}  &  15.78  &  33.04      \\
\hspace{0.3cm} \textbf{$\beta_{\operatorname{random}}$=6}
&  2.73  &  8.10  &  41.72  &  26.32  &  \bf{37.35}      \\
\hspace{0.3cm} \textbf{$\beta_{\operatorname{random}}$=9}
&  2.17  &  8.83  &  37.83  &  30.27  &  36.03      \\
\hspace{0.3cm} \textbf{$\beta_{\operatorname{random}}$=12}
&  1.75  &  \bf{9.45}  &  32.27  &  \bf{32.59}  &  32.33      \\

\textbf{MixEdit} (Ours)  &  \bf{3.95}  &  8.52   &  \bf{47.81}  &  18.31  &  36.15      \\

\bottomrule
\end{tabular}}

\caption{
Affinity and Diversity of data augmentation methods.
All the models are trained using realistic or pseudo Chinese HSK dataset.
}
\label{tab:affinity_diversity_hsk}

\end{table}
\begin{table}[thbp!]
\centering
\scalebox{0.67}{
\begin{tabular}{lllcccc}
\toprule

\bf{Method}  &  \bf{Data}  &  \bf{Aff.}$^\uparrow$  &  \bf{Div.}  &
\bf{P}  &  \bf{R} & $\mbox{\bf{F}}_{0.5}$           \\ 
\hline

\bf{Baseline}  &  124K  &  -     &  8.78   &  56.03  &  37.60  &  51.03     \\
\bf{DN}  &  249K  &  0.41  &  10.70  &  53.89  &  39.72  &  50.30     \\
\bf{PN}   \\
\bf{RT}         
&  249K  &  0.75  &  10.54  &  50.32  &  39.82  &  47.80      \\

\hspace{0.3cm} \bf{Round=1}  
&  249K  &  1.91  &  7.46  &  56.64  &  36.75  &  51.11     \\
\hspace{0.3cm} \bf{Round=2}  
&  249K  &  \bf{1.93}  &  7.56  &  56.07  &  40.15  &  \bf{51.95}     \\
\hspace{0.3cm} \bf{Round=4}  
&  249K  &  1.90  &  7.70  &  53.88  &  43.04  &  51.30     \\
\hspace{0.3cm} \bf{Round=8}  
&  249K  &  1.77  &  7.92  &  54.03  &  41.32  &  50.90     \\
\hspace{0.3cm} \bf{Round=16} 
&  249K  &  1.59  &  8.24  &  54.43  &  40.88  &  51.04     \\

\bf{BT}   \\
\hspace{0.3cm} \bf{$\beta_{\operatorname{random}}$=0}
&  249K  &  0.65  &  6.96  &  56.46  &  33.62  &  49.71      \\
\hspace{0.3cm} \bf{$\beta_{\operatorname{random}}$=3}
&  249K  &  1.47  &  7.67  &  54.94  &  42.87  &  52.01      \\
\hspace{0.3cm} \bf{$\beta_{\operatorname{random}}$=6}
&  249K  &  \bf{1.57}  &  8.22  &  54.04  &  46.38  &  \bf{52.31}      \\
\hspace{0.3cm} \bf{$\beta_{\operatorname{random}}$=9}
&  249K  &  1.53  &  8.59  &  53.63  &  45.76  &  51.84      \\
\hspace{0.3cm} \bf{$\beta_{\operatorname{random}}$=12}
&  249K  &  1.45  &  8.84  &  53.79  &  43.03  &  51.22      \\

\bf{MixEdit (Static)}   
&  249K  &  \bf{2.33}  &  8.52   &  57.98  &  37.78  &  \bf{52.38}      \\
\bf{MixEdit (Dynamic)}   
&  -     &  -     &  -      &  57.96  &  39.00  &  \bf{52.83}      \\
\bottomrule
\end{tabular}}

\caption{
Results on the combination of the realistic and pseudo BEA-train datasets for various data augmentation methods.
\textbf{MixEdit (Dynamic)} generate pseudo data on-the-fly during fine-tuning.
}
\label{tab:combination}
\end{table}

We conduct extra experiments illustrated in Section~\ref{subsec:analysis} on English CLang-8 and Chinese HSK datasets.
The results are listed in Table~\ref{tab:affinity_diversity_clang8} and Table~\ref{tab:affinity_diversity_hsk}, respectively.
Surprisingly, backtranslation (BT) achieves the highest F0.5 on Chinese HSK.
We speculate that Chinese grammatical errors are inherently more intricate, providing an advantage for BT as it can generate pseudo grammatical errors that are closer to authentic ones.
Nonetheless, it is worth noting that BT relies on an additional model to generate grammatical errors, which introduces efficiency concerns.
The Pearson's correlation coefficients between F${0.5}$ and Affinity are 0.6239 on CLang8, and 0.7717 on HSK.
The results indicate a strong or moderate correlation between Affinity and F${0.5}$ on different datasets, demonstrating the generalization of our proposed approach.

\subsection{Complementary Effectiveness of Pseudo Data}
\label{app:complement_psuedo_data}

We have analyzed the effectiveness of GEC data augmentation methods from the lens of Affinity and Diversity measures in Section~\ref{subsec:analysis}.
However, one may argue that MixEdit gains unfair advantages since the information about the positions of grammatical errors is only visible to MixEdit.
To this end, we investigate the complementary effectiveness of pseudo data in this section.
Specifically, we construct a combination dataset by appending the realistic data of BEA-train to the pseudo data generated by data augmentation methods, where a target sentence correspond to two source sentences.

We train GEC models using the combination dataset and report the results in Table~\ref{tab:combination}.
Similarly, DN and RT perform worse than the baseline since these methods inject considerable undesired noise, which makes the model prone to inaccurate corrections.
PN arrives its peek of F$_{0.5}$ score at Round=2, rather than Round=8 in Table~\ref{tab:affinity_diversity_bea}.
BT arrives the peek at $\beta_{\operatorname{random}}=6$, and most selections of $\beta_{\operatorname{random}}$ can improve GEC models.
MixEdit (Static) performs the best among all static data augmentation methods.
Furthermore, MixEdit (Dynamic) achieves the highest F$_{0.5}$ score, demonstrating the effectiveness of dynamic pseudo data construction.

\subsection{Incorporating PN into MixEdit}\label{app:pn+mixedit}

\begin{table}[tp!]
\centering
\scalebox{0.78}{
\begin{tabular}{lcc}
\toprule

&  \textbf{CoNLL-14-\textit{test}}  &  \textbf{BEA-19-\textit{test}} \\ 

\textbf{Corruption Ratio}  & \textbf{P}/\textbf{R}/\textbf{$\mbox{\textbf{F}}_{0.5}$} & \textbf{P}/\textbf{R}/\textbf{$\mbox{\textbf{F}}_{0.5}$} \\

\hline

0.00
&  76.81/45.00/67.30  &  76.37/62.71/\textbf{73.18}      \\

0.02
&  74.94/48.70/67.65  &  75.14/64.04/72.62      \\

0.05 
&  74.70/49.73/\textbf{67.88}  &  74.78/64.67/72.51      \\

0.10
&  72.42/50.68/66.70  &  73.70/65.25/71.84      \\

0.15
&  73.26/48.95/66.64  &  72.96/64.46/71.08      \\

\bottomrule
\end{tabular}}
\caption{
Results of incorporating PN into MixEdit.
The Corruption Ratio indicates the probability of adding grammatical errors to a token in the source sentence.
}
\label{tab:pn+mixedit}
\end{table}

We also explore the effectiveness of incorporating PN into MixEdit in a dynamic manner.
In this approach, we applying PN after running MixEdit.
This means that we randomly add grammatical errors to the pseudo source generated by MixEdit.
We report the results of varying PN corruption ratios in Table~\ref{tab:pn+mixedit}.
Our findings suggest that when the correction ratio is low, PN can benefit GEC models on CoNLL-14-\textit{test}.
However, PN decreases the F$_{0.5}$ scores on BEA-19-\textit{test} regardless of different correction ratios.
We attribute this to the different characteristics of the two evaluation datasets, where CoNLL-14 contains more typical grammatical errors written by language learners.

\end{document}